\documentclass[conference]{IEEEtran}

\IEEEoverridecommandlockouts                              

\usepackage{graphics} 
\usepackage{epsfig} 
\usepackage{mathptmx} 
\usepackage{times} 
\usepackage{amsmath} 
\usepackage{amssymb}  
\usepackage{graphicx}
\usepackage{array}
\usepackage{booktabs}
\usepackage{dialogue}
\usepackage{multirow}
\usepackage{color}
\usepackage[normalem]{ulem}
\usepackage{xcolor}
\usepackage{listings}
\usepackage{caption}
\usepackage{subfig}
\usepackage{cite}
\usepackage[colorlinks=true, linkcolor=blue, citecolor=blue, urlcolor=blue]{hyperref}
\usepackage{url}
\usepackage{float}
\usepackage{placeins}

\newcolumntype{R}[1]{>{\raggedleft\arraybackslash}p{#1}}

\begin{document}

\title{\LARGE \bf
Benchmarking Vision, Language, \& Action Models in Procedurally Generated, Open-Ended Action Environments
\thanks{* equal contribution, alphabetical order.}
\thanks{+ Corresponding Author: harshsikka@gatech.edu}
}

\author{\IEEEauthorblockN{Pranav Guruprasad$^{*+12}$, Yangyue Wang$^{*12}$, Sudipta Chowdhury$^{*1}$, Harshvardhan Sikka$^{123}$, Paul Pu Liang$^{4}$}\\
\newline
\IEEEauthorblockA{$^1$Manifold Research}
\newline
\IEEEauthorblockA{$^2$Metarch.ai}
\newline
\IEEEauthorblockA{$^3$Georgia Tech}
\newline
\IEEEauthorblockA{$^4$MIT}
}

\maketitle
\thispagestyle{empty}
\pagestyle{empty}

\begin{abstract}
Vision-language-action (VLA) models represent an important step toward general-purpose robotic systems by integrating visual perception, language understanding, and action execution. However, systematic evaluation of these models, particularly their zero-shot generalization capabilities in procedurally out-of-distribution (OOD) environments, remains limited. In this paper, we introduce MultiNet v0.2, a comprehensive benchmark designed to evaluate and analyze the generalization performance of state-of-the-art VLMs and VLAs—including GPT-4o, GPT-4.1, OpenVLA, Pi0 Base, and Pi0 FAST—on diverse procedural tasks from the Procgen benchmark. Our analysis reveals several critical insights: (1) all evaluated models exhibit significant limitations in zero-shot generalization to OOD tasks, with performance heavily influenced by factors such as action representation and task complexity; (2) VLAs generally outperforms other models due to their robust architectural design; and (3) VLM variants demonstrate substantial improvements when constrained appropriately, highlighting the sensitivity of model performance to precise prompt engineering. We release our benchmark, evaluation framework, and findings to enable the assessment of future VLA models and
identify critical areas for improvement in their application to out-of-distribution digital tasks. 
\end{abstract}


\section{Introduction}

Recent advancements in large-scale vision-language models (VLMs) and vision-language-action models (VLAs) have demonstrated remarkable capabilities across various domains, including image recognition, natural language understanding, multimodal association, and preliminary robotics applications \cite{gato2022, openvla2024, rt22023}. These models promise a future where general-purpose AI systems can interpret visual inputs, comprehend language commands, and execute appropriate actions in diverse scenarios.

However, a significant challenge remains: ensuring these models can generalize effectively to out-of-distribution (OOD) tasks. Current models often struggle with zero-shot transfer capabilities, particularly when confronted with novel scenarios or tasks that differ substantially from their training data \cite{benchmarking2024}. This limitation is especially evident in procedurally generated environments, such as those provided by the Procgen benchmark\cite{cobbe2020leveragingproceduralgenerationbenchmark}, which are designed to test visual understanding, decision-making, and action generation capabilities in varied and unpredictable settings. 

The architecture and training methodologies of these models play a crucial role in their generalization abilities. Factors such as the nature of the action space (continuous vs. discrete), the domain of training data (real-world robotics vs. simulated environments), and the methods of input-output processing can significantly impact performance. For instance, models trained predominantly on robotics data may not perform well in simulated game environments due to differences in action representation and environmental complexity.

To address these challenges, we introduce \textbf{MultiNet v0.2}, a comprehensive benchmarking effort aimed at evaluating the generalist capabilities of VLMs and VLAs on procedurally generated tasks. Our study encompasses a diverse set of models, including GPT-4o, GPT-4.1, OpenVLA, Pi0 Base, and Pi0 FAST, assessed across multiple Procgen datasets\cite{openai_gpt4, openai_gpt41,openvla2024 ,pi02024, pi0fast2025, cobbe2020leveragingproceduralgenerationbenchmark}. 

Our primary contributions in this paper are:
\begin{itemize}
    \item A systematic benchmarking framework that evaluates the performance of state-of-the-art models on a variety of procedurally generated game environments.
    \item Detailed profiling results for an initial set of SoTA VLMs and VLAs, along with analysis of their performance.
    \item Analysis of the impact of architectural choices, training data, and output processing techniques on model generalization, highlighting the limitations of current approaches.
    \item Insights into how factors like action space representation and image complexity influence model performance.
\end{itemize}

The remainder of this paper is structured as follows: Section 2 reviews related work in the field; Section 3 details our experimental setup and datasets; Section 4 presents our empirical results and analysis; and Section 5 discusses the implications of our findings and outlines directions for future research.

\begin{figure}
    \centering
    \includegraphics[width=\linewidth]{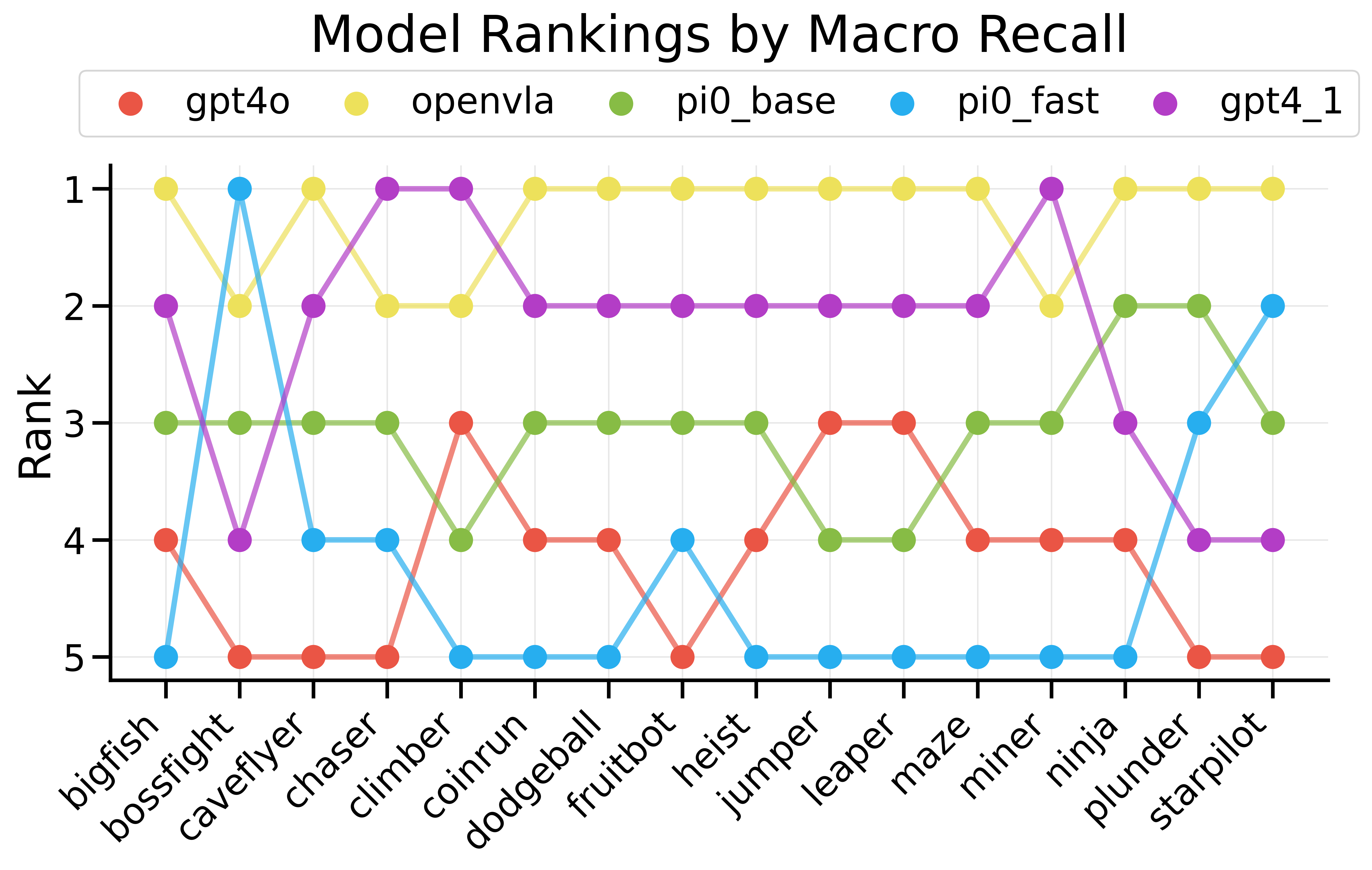}
    \caption{Model rankings across datasets by macro recall, highlighting consistent top performance by OpenVLA and GPT-4.1. }
    \label{fig:fig10}
\end{figure}

\section{Related Work}
Our work builds on the rapid advancement in vision-language-action (VLA) models for embodied AI, both in physical robots and digital environments, that enable agents to perceive, reason, and act across diverse tasks and embodiments. In this section, we review the landscape of related work, organized into two complementary areas: \emph{evaluation in digital environments} and \emph{evaluation of generalist VLAs}. For comprehensive surveys on VLAs and computer use agents, see \cite{OSAgentSurvey2024, RoboVLM2024, VLASurvey2024, ComputerUseAgentSurvey2025}.

\textbf{Evaluation in digital environments}. Digital benchmarks play a crucial role in measuring agents' decision-making, perception, and planning skills in simulated settings. We categorize prominent digital benchmarks into three groups:
\begin{itemize}
    \item \textbf{Gameplay Benchmarks}. Benchmarks such as Gymnasium, which wraps the Arcade Learning Environment (ALE) for Atari games and provides continuous-control tasks, have become standard testbeds for reinforcement-learning agents\cite{ALE2012}. More recent procedurally generated game benchmarks such as Procgen, MineDojo and competition and crafting game benchmarks such as AlphaStar,  Crafter introduce a family of tasks that probe models' ability in 2D/3D games with more complex objectives and environments to further assess agents' ability to adapt and generalize\cite{cobbe2020leveragingproceduralgenerationbenchmark, fan2022minedojo, vinyals2019grandmaster, Crafter2022}. 
    \item \textbf{Interface and Web-Interaction Benchmarks}. As agent evaluation transitions from controlled gameplay simulations to more general computer use tasks, benchmarks such as Mind2Web, AITW, and OSWorld aim to evaluate proficiency in tasks such as web browsering, menu navigation, and form completion across environments such as web applications, Windows, MacOS, Android, and Linux. \cite{Mind2Web2023, AITW2023, OSWorld2024, GUIWorld2024, VideoGUI2024}. 
\end{itemize}
In recent years, agents driven by large language models have been evaluated in subsets of these benchmarks\cite{Cradle2024, VOYAGER2023}.

We extend digital evaluation by systematically including state-of-the-art VLAs alongside VLM-based agents, applying a unified metric suite across all procedurally generated and discrete-control environments.

\textbf{Evaluation of generalist VLAs}. Generalist VLAs are commonly evaluated using in-domain robotics benchmarks such as OpenX, VLABench, and LIBERO, consisting of common physical manipulation tasks and robot embodiments \cite{openx2024, bridgev2_2023, VLABench, AgiBot_GO1, 3DAffordanceNet2021}. These assessments focus on metrics such as task completion rates, sim-to-real transfer fidelity, and robustness to sensor noise. Some recent efforts such as Magma \cite{Magma2025} have attempted to narrow cross-domain transfer, but none systematically unify digital and physical benchmarks under consistent metrics and datasets.

In MultiNet v0.2, we repurpose robotics-trained VLAs and general-purpose VLMs as digital agents and benchmark them in procedurally generated 2D environments.

\section{Experimental Setup} 

\subsection{Dataset}
We utilized offline trajectories from expert reinforcement learning (RL) agents trained on the Procgen dataset, which were sourced from Facebook’s publicly available repository \footnote{https://dl.fbaipublicfiles.com/DGRL/1M/expert/} \cite{cobbe2020leveragingproceduralgenerationbenchmark}. To facilitate consistent and straightforward data handling, we translated these trajectories into the TensorFlow Datasets (TFDS) format\footnote{https://www.tensorflow.org/api\_docs/python/tf/data/Dataset}. This standardized format ensured streamlined loading and usage across all subsequent profiling experiments. 

For evaluation purposes, we established test splits comprising 10\% of the episodes randomly sampled from each subdataset. Given that the Procgen dataset consists of 16 distinct subdatasets, this approach resulted in balanced and representative test splits for comprehensive evaluation. Importantly, all profiling and performance assessment experiments reported in this study were conducted exclusively on these designated test splits, ensuring that our results reflect the models' genuine generalization capabilities without contamination from training data.

\subsection{Models}
In this study, we leveraged state-of-the-art Vision-Language-Action (VLA) models with predefined full-sized weights to evaluate performance comprehensively. The models and their respective sources were:

OpenVLA: The weights were obtained from the HuggingFace repository under the identifier openvla/openvla-7b.\footnote{https://huggingface.co/openvla/openvla-7b}

Pi0 Base: We used weights stored at s3://openpi-assets/checkpoints/pi0{\_}base/params. We obtained this path from the Openpi open-source codebase\footnote{https://github.com/Physical-Intelligence/openpi}

Pi0 Fast: We used weights accessible via s3://openpi-assets/checkpoints/pi0{\_}fast{\_}base/params. We obtained this path from the Openpi open-source codebase.

All models were employed in their complete forms, without any quantization techniques, to ensure accurate and undiluted performance evaluation.

To align the models with the Procgen task context and dataset specifics, we performed targeted adaptations during ingestion and inference.

\subsubsection{Data Ingestion}

\textbf{GPT 4o and GPT 4.1:} Employed the Genesis prompt engineering framework, which acts as a crucial intermediary between the data and the model, by translating the raw Procgen trajectory data (images and associated metadata) into a rich, structured textual representation that primes the VLM for analyzable outputs. Further information about the Genesis framework is detailed in Appendix Section VII-C. 

\textbf{OpenVLA:} Received visual observations per timestep as direct image inputs, accompanied by succinct single-line text prompts describing the corresponding subdataset task.

\textbf{Pi0:} Incorporated the first image view as the primary visual observation, supplemented with zero arrays for the second and third image views, with image masks indicating True for the primary view and False for others. Proprioceptive states were also set as zero arrays. Each timestep included a concise single-line task description prompt.

\textbf{Pi0 with Fast:} Similar to Pi0 Base, Pi0 Fast utilized the first image view as its primary visual observation but employed zero arrays for the second and third image views with image masks set to True for all three views. Zero arrays represented the proprioceptive states, and each timestep included a single-line task description prompt.

\subsubsection{Model Adaptation}

\textbf{OpenVLA:} Adaptation involved restricting the autoregressive step to one, ensuring single-dimensional action vector predictions. Generated actions were unnormalized using statistics computed across the entire Procgen subdataset and subsequently rounded to discrete actions. To compute evaluation metrics such as Brier MAE  \ref{eq:1}, we first extracted logits from the Llama 2 backbone and computed the corresponding probabilities \cite{touvron2023llama2openfoundation}. We then initialized and cached mappings from Llama vocabulary tokens to Procgen integer actions. During inference, this approach allowed us to group the probabilities of each vocabulary token according to their associated integer action classes after unnormalization, while maintaining efficient inference throughout the process.

\textbf{Pi0 Base:} The action horizon was constrained to one timestep, with flow matching denoising executed over ten steps (default configuration). Actions were predicted with a default dimension of 32. The first dimension of the predicted action was picked to ensure a final single-dimensional action vector prediction, which was then unnormalized based on Procgen dataset-wide statistics (same dataset statistics used for OpenVLA), and discretized by rounding.

\textbf{Pi0 Fast:} Adaptation entailed adjusting the action horizon to one timestep and setting the action dimension to one, compatible with Procgen’s single-dimensional action space. To explicitly generate the correct action value, the decoding steps were limited to four tokens: ``Action'', ``:'', a space character, and the Paligemma (VLM backbone of Pi0 models) location token. To compute probabilities for Brier MAE, leveraging the static nature of the BPE tokenizer,  token mappings between the Paligemma token IDs and Procgen integer actions were established. During inference, probabilities extracted from the Paligemma backbone logits were aggregated according to this mapping. Recognizing Pi0 Fast's significantly slower inference speed (approximately ten times slower than OpenVLA), we optimized runtime by caching embeddings for the two static zero-images after initial processing by the SigLIP embedding \cite{zhai2023sigmoidlosslanguageimage}, effectively doubling inference speed.

\subsection{Inference Infrastructure}

For systematic performance evaluation and model profiling, we utilized dedicated hardware resources optimized for each model's computational requirements:

\begin{itemize}
  \item OpenVLA inference was executed on an NVIDIA L4 GPU instance, selected for its balance of computational efficiency and memory capacity, ideal for models of intermediate complexity.

  \item Pi0 Base utilized a single NVIDIA A100 GPU instance equipped with 40 GB of memory, providing ample computational power and memory bandwidth for accurate and efficient inference.
  \item Pi0 Fast, due to its greater computational demands and inference complexity, was allocated four NVIDIA A100 GPU instances, each with 40 GB of memory, facilitating parallel processing to substantially enhance inference throughput.
  \item In contrast, GPT 4x inference was managed externally through OpenAI's Batch API, eliminating local processing overhead and leveraging scalable cloud resources to efficiently handle its computational requirements.

\end{itemize}

\subsection{Evaluation metrics}

In order to capture the ability of state-of-the-art VLAs and VLMs to generalize to completely unseen, OOD action data, we carefully curated a list of metrics to evaluate the models as fairly as possible. These metrics address diverse aspects of model performance, such as how well calibrated they are in their estimates, reliance on outliers, how they are affected by class imbalance, how well they are able to restrict their predictions to the valid action space, whether the models are biased towards certain kinds of subdatasets or actions, etc.

\begin{itemize}
    \item \textbf{Brier Mean Absolute Error}
    \begin{equation}
    \text{Brier MAE} = \frac{1}{N} \sum_{t=1}^{N} \sum_{i=1}^{R} |f_{ti} - o_{ti}| \label{eq:1}
    \end{equation}

    where $f_{ti}$ is the probability of the model prediction for class i at timestep t, $o_{ti}$ is the $i^{th}$ dimension of the one hot representation of the ground truth action for the same timestep, R is the possible number of classes, and N is the total number of timesteps.

    Brier MAE is a variation of the original Brier score \cite{Brier1950}, which is a useful method to measure the accuracy of probabilistic predictions. In our case, for a given timestep and the associated action space, each model computes a probability for each action that represents how likely it is to be equal to the truth label. This measures how well the predictions of the models are calibrated. Brier score penalizes predictions that are both wrong and confident more severely than those that are wrong but uncertain. Additionally, unlike Mean Squared Error, using Mean Absolute Error maintains the original scale of the data, making it potentially more interpretable in the action space's context. MAE is also less sensitive to outlier predictions than squared error metrics, thus giving a more stable evaluation of typical performance. The maximum possible Brier score is 2, which is assigned to invalid predictions as a high penalty.

    \item\textbf{Normalized Brier Mean Absolute Error}
    Average of Brier Absolute Errors that have been min-max normalized using the minimum and maximum Brier absolute errors over all timesteps. A discrepancy in trends between regular Brier MAE and Normalized Brier MAE indicates reliance on outlier predictions that lead to abnormally high or low Brier MAEs.

    \item\textbf{Normalized Quantile Filtered Brier Mean Absolute Error}
    This metric value is obtained by filtering out the Brier Absolute Errors that are lesser than or equal to the 5th percentile error or greater than or equal to the 95th percentile error, and then normalizing based on the quantile filtered minimum and maximum errors. This metric helps indicate how good or bad the majority of the predictions are. Discrepancy in trends between normalized and normalized quantile filtered Brier MAEs indicates the reliance on a few extremely good or extremely bad predictions.

    \item\textbf{Maximum relative Brier Mean Absolute Error}
    \begin{equation}
        \text{Max relative Brier MAE} = \frac{\max(MAE)}{\text{median}(MAE)}
    \end{equation}

    This metric quantifies how much the worst-case error deviates from the typical (median) error produced by the model for a given subdataset of Procgen. A value significantly greater than 1 indicates the presence of some extremely bad predictions that can skew the results.

    \item\textbf{Micro Precision, Recall, F1 score, Exact Match rate}
    \begin{equation}
        \text{Micro Precision} = \frac{\text{True Positives}}{\text{True Positives + False Positives}}
    \end{equation}

    \begin{equation}
        \text{Micro Recall} = \frac{\text{True Positives}}{\text{True Positives + False Negatives}}
    \end{equation}

    \begin{equation}
        \text{F1 score} = \frac{\text{2x(Precision x Recall)}}{\text{Precision + Recall}}
    \end{equation}

    \small
    \begin{equation}
        \text{Exact Match Rate} = 
        \frac{\text{Num timesteps where preds = labels}}{\text{Total number of timesteps}}
    \end{equation}
    \normalsize
    
    For our case, where it is a multi-class, single correct prediction task and the classes (actions) are mutually exclusive, false positives are equal to false negatives. Simply, an incorrectly predicted action is a false positive for its class, while also a false negative for the ground truth action class. Due to this, precision = recall = f1 score = exact match rate when calculating micro metrics in our case. The micro precision/recall/f1 is calculated over all timesteps of the test split of a given subdataset. This metric measures how precise the models are - high precision indicates that the model gets a high number of predictions correct, and a low precision suggests the model is frequently predicting actions that don't match the ground truth.

    \item\textbf{Percentage invalids}
    Predictions that fall outside the valid action space of a given subdataset of Procgen are considered invalid. In the case of the VLMs - GPT 4o, and GPT 4.1, the outputs that are not in the desired format of a vector the size of the action space with discrete action keys and their probabilities summing to 1 as its elements, are also considered invalid. Thus, percentage invalids is calculated as

    \small
    \begin{equation}
        \text{Percentage invalids} = \frac{\text{Timesteps with invalid preds}}{\text{Total number of timesteps}}*100
    \end{equation}
    \normalsize

    A high invalid percentage indicates that the model struggles to produce outputs in the target action space.

     \item\textbf{Micro Precision without Invalids}
     This metric is a variation of micro precision where the false positives are only counted for valid predictions. By not considering the invalid predictions, this metric provides more insight into the model's performance in the cases where it is able to produce valid outputs. A discrepancy in the trends between micro precision and micro precision without invalids indicates that a model would be more capable if it were fine-tuned, prompted, or had its output processed to constrain the predictions to the valid action space.

     \item\textbf{Class-wise Precision, Recall, F1 scores}
     Class-wise variations of the precision, recall, and F1 metrics are the same metrics calculated for each action class within the valid action space of a subdataset individually. The class-wise precision, recall, and F1 scores of a model would be calculated for a given action class x subdataset combination. Average class-wise metrics for a model are calculated by averaging the class-wise x subdataset metric values over all subdatasets to report a unified set of class-wise metrics for a given model.

     These metrics help to understand model biases and preferences towards specific action classes.

     \item\textbf{Macro Precision, Recall, and F1 scores}

     \begin{equation}
        \text{Macro Precision} = \frac{1}{K} \sum_{k=1}^{K} \frac{TP_k}{TP_k + FP_k}
     \end{equation}

     \begin{equation}
        \text{Macro Recall} = \frac{1}{K} \sum_{k=1}^{K} \frac{TP_k}{TP_k + FN_k}
     \end{equation}
     \begin{equation}
        \text{Macro F1} = \frac{1}{K} \sum_{k=1}^{K} \frac{2 \cdot \text{Precision}_k \cdot \text{Recall}_k}{\text{Precision}_k + \text{Recall}_k}
     \end{equation}
     K here refers to the number of classes for a given subdataset. Macro precision, recall, and F1 metrics are calculated by averaging the class-wise metrics across all classes in the valid action space of a given subdataset for a given model. These metrics are crucial, as they help alleviate the effects of the dominant majority class that skews performance. With micro metrics, the majority classes are favored due to global aggregation. This leads to a biased view of performance, especially in the Procgen expert data, as there are numerous cases of class imbalance, which can skew the values of the micro metrics. Macro metrics prove to be a better indicator of a model's performance when compared to micro metrics, irrespective of the nature of the test dataset, as they treat all classes equally regardless of size. Within this, Macro recall is typically the most representative metric for performance as it is neither affected by the majority classes, as in the case of micro metrics, nor is it affected by the performance of models on the rare classes, as in the case of macro precision or F1 scores.

\end{itemize}

\section{Results and Discussion}

\subsection{Models Exhibit Poor Zero-Shot Out-of-Distribution (OOD) Generalization}
\subsubsection{Comparing General Performance} In evaluating the generalization capability of GPT 4o, GPT 4.1, OpenVLA, Pi0 Base, and Pi0 FAST, we observed universally poor performance on zero-shot out-of-distribution (OOD) tasks. The analysis using various performance metrics underscored significant limitations across these models.

\begin{figure}[h]
	\centering
	\includegraphics[width=9cm]{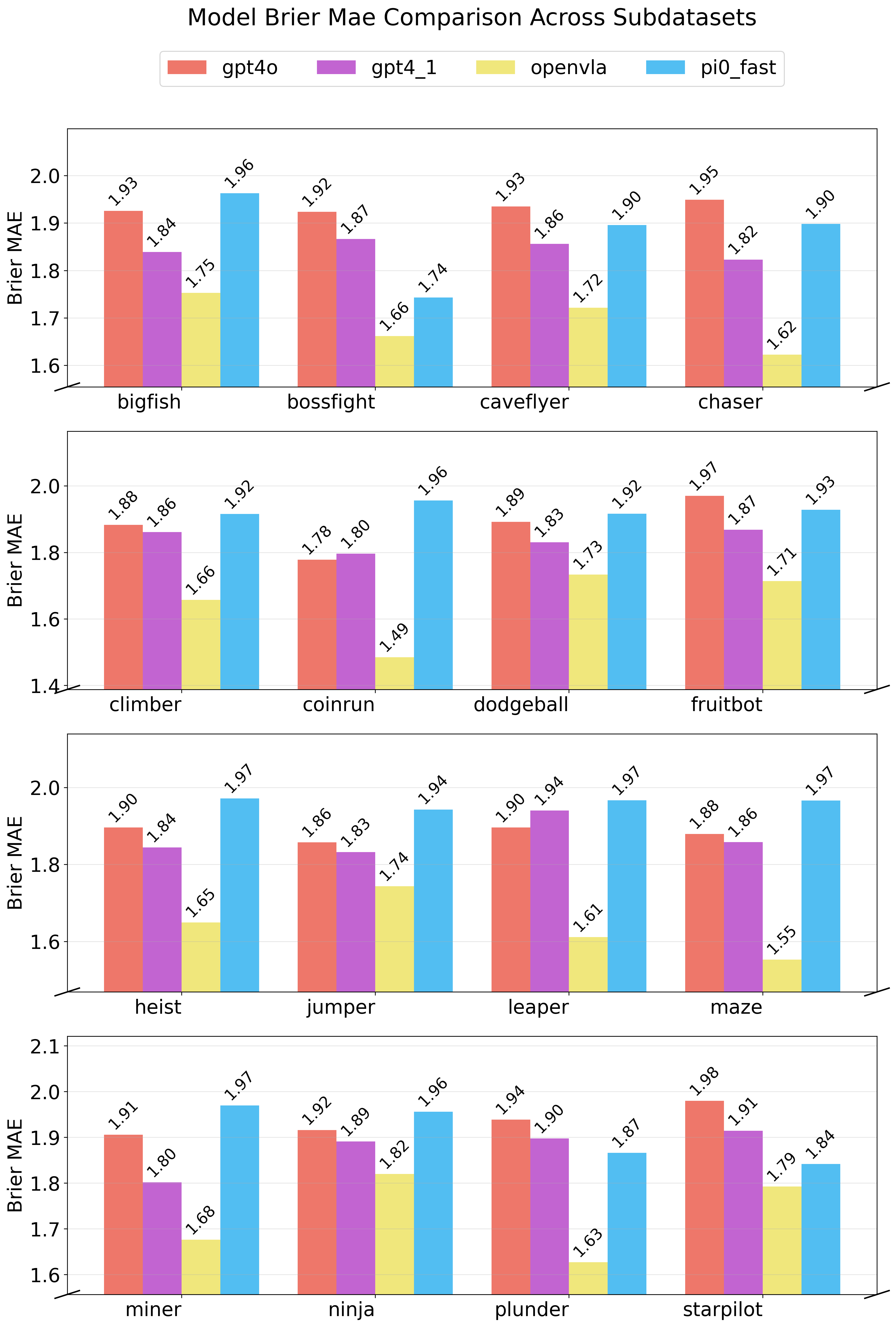}
	\caption{Brier Mean Absolute Error scores across 4 models - GPT 4o, OpenVLA, GPT 4.1, and Pi0 FAST. Pi0 Base is a diffusion-based model and can not be evaluated using Brier MAE due to a lack of logits in its inference architecture. All models display Brier MAE close to 2, indicating poor performance.}
	\label{fig:fig1}
\end{figure}

As seen in Figure \ref{fig:fig1}, the Brier Mean Absolute Error (MAE) scores revealed poor probability calibration, as the scores approached the maximum value of 2 across all models and datasets. Specifically, GPT 4o, GPT 4.1, and Pi0 FAST consistently scored above 1.70, while OpenVLA generally exceeded 1.50. The subdatasets on which models exhibited the weakest performance were Starpilot for GPT 4o, Leaper for GPT 4.1, Miner for Pi0 FAST, and Ninja for OpenVLA.

\begin{figure}[h]
	\centering
	\includegraphics[width=9cm]{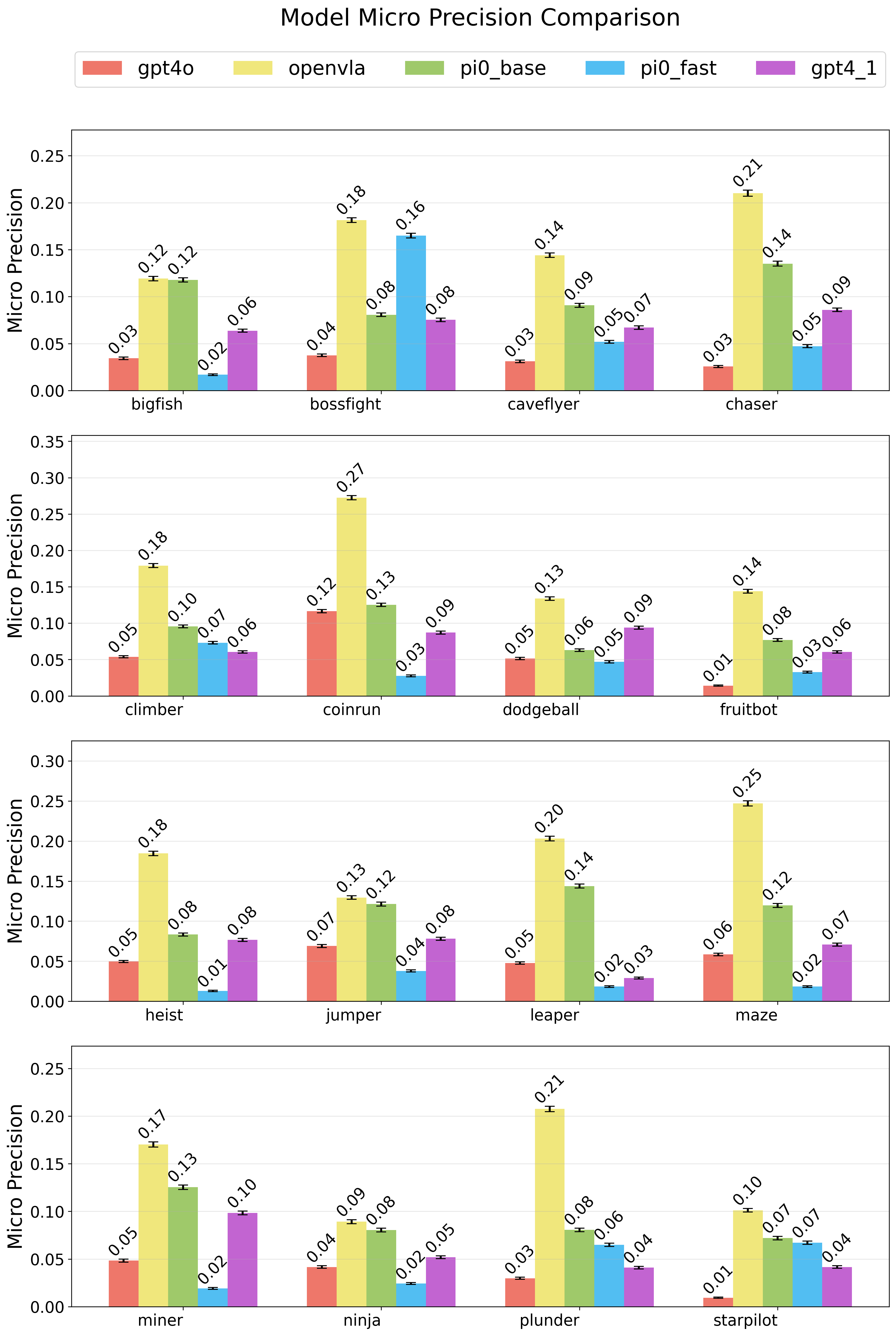}
	\caption{Micro precision across all 5 models - GPT 4o, OpenVLA, Pi0 Base, Pi0 FAST, and GPT 4.1. When considering all 16 subdatasets, Pi0 FAST overall showcases the worst performance, and OpenVLA the best performance.   }
	\label{fig:fig2}
\end{figure}

Analyzing micro-level metrics, precision values were low across all models and subdatasets. OpenVLA achieved the highest micro precision at approximately 27\% on the Coinrun dataset, whereas GPT 4o showed the lowest precision at 1\% on Starpilot. Pi0 FAST reached a minimal precision of around 1\% on Heist. Overall, as can be inferred from Figure \ref{fig:fig2}, the performance hierarchy based on micro precision was: OpenVLA $>$ Pi0 Base $>$ GPT 4.1 $>$ GPT 4o $>$ Pi0 FAST.

\begin{figure}
    \centering
    \includegraphics[width=9cm]{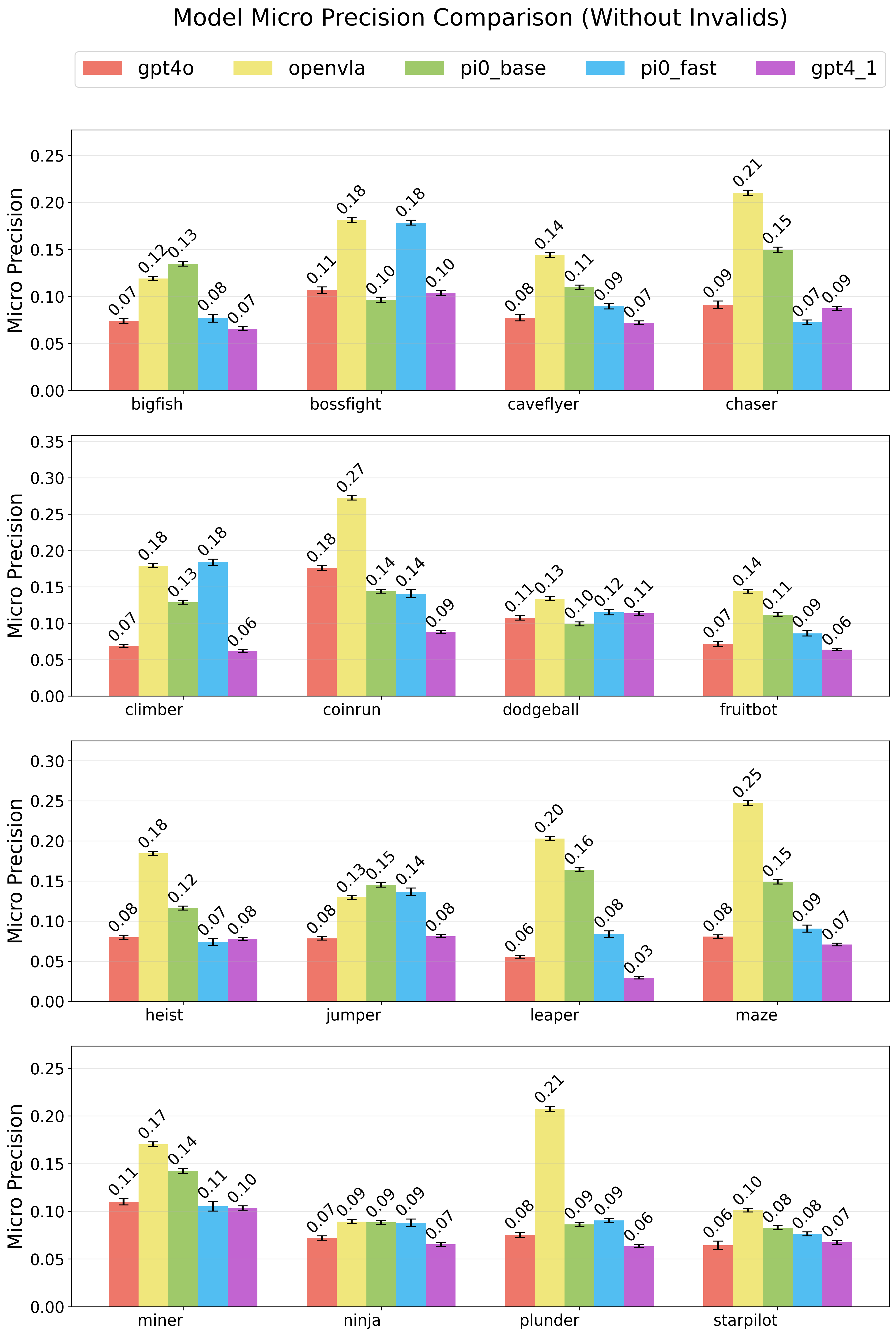}
    \caption{Micro precision across all 5 models - GPT 4o, OpenVLA, Pi0 Base, Pi0 FAST, and GPT 4.1 without invalids. When considering all 16 subdatasets, GPT 4.1 overall showcases the worst performance, and OpenVLA the best performance.}
    \label{fig:fig3}
\end{figure}

When considering micro precision without invalid predictions, as seen in Figure \ref{fig:fig3}, significant improvements were noted for GPT 4o and Pi0 FAST, highlighting their potential if constrained or further fine-tuned to produce valid action outputs. GPT 4o's precision notably improved, surpassing Pi0 Base in a few subdatasets such as Coinrun, Bossfight, and Dodgeball. Nevertheless, the adjusted performance order still positioned OpenVLA at the top, followed by Pi0 Base, Pi0 FAST, GPT 4o, and GPT 4.1.

\begin{figure}
    \centering
    \includegraphics[width=9cm]{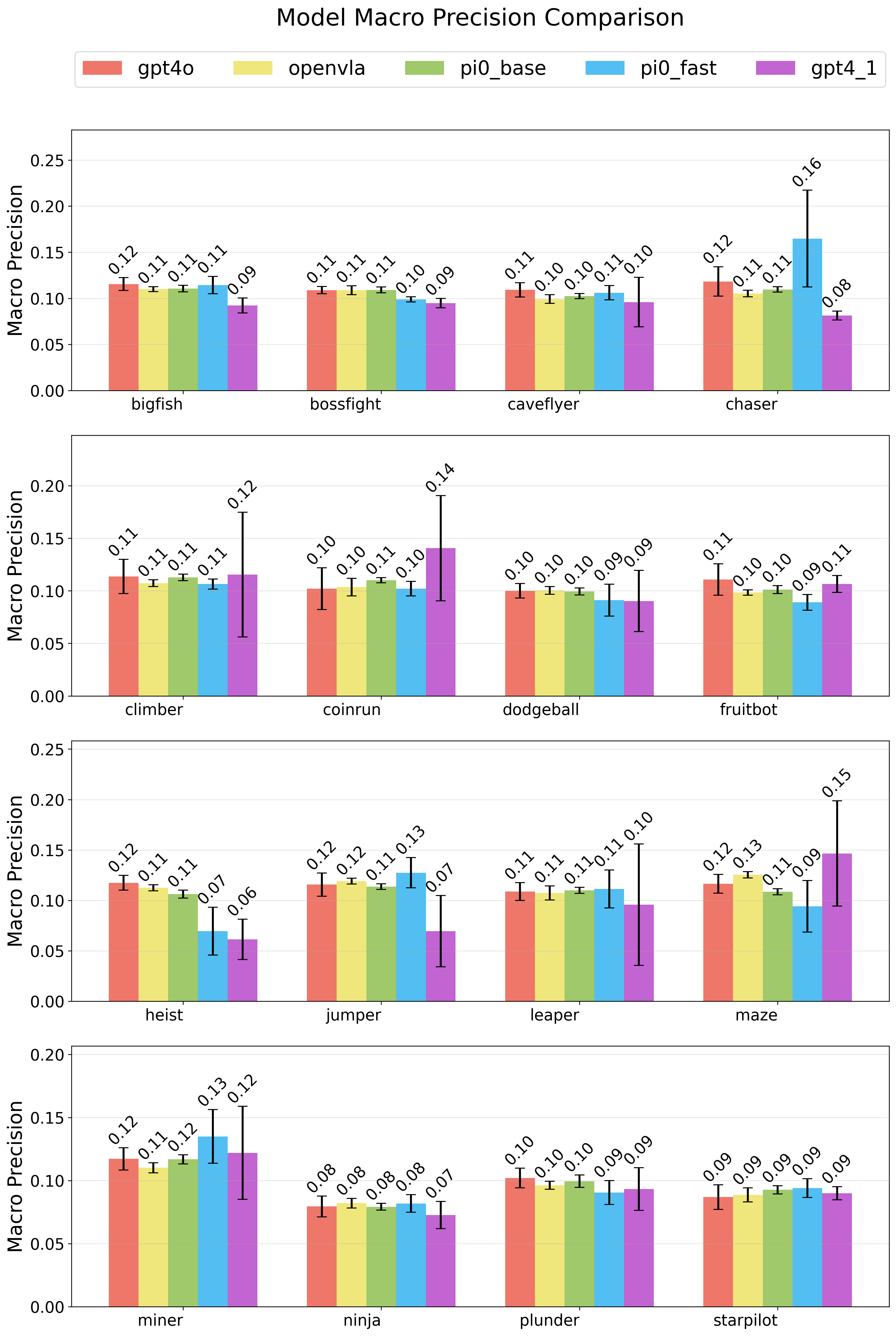}
\caption{Macro precision across all 5 models. GPT 4.1 and Pi0 FAST show variability and preference towards specific subdatasets. OpenVLA shows reliability on the majority classes, whereas GPT 4o displays relatively stronger performance on minority classes.}
	\label{fig:fig4}
\end{figure}

Macro precision analysis revealed that GPT 4o, OpenVLA, and Pi0 Base maintained similar performance levels across most datasets. As seen in Figure \ref{fig:fig4}, conversely, GPT 4.1 and Pi0 FAST demonstrated significant variability, implying distinct preferences for specific task types. Despite these observations, macro precision remained generally low across all models, with a maximum slightly above 15\%. Importantly, comparing its macro and micro precision performances, it is clear that OpenVLA's performance heavily favored the majority classes, while GPT 4o exhibited relatively stronger performance on minority classes.

\begin{figure}[h]
	\centering
	\includegraphics[width=9cm]{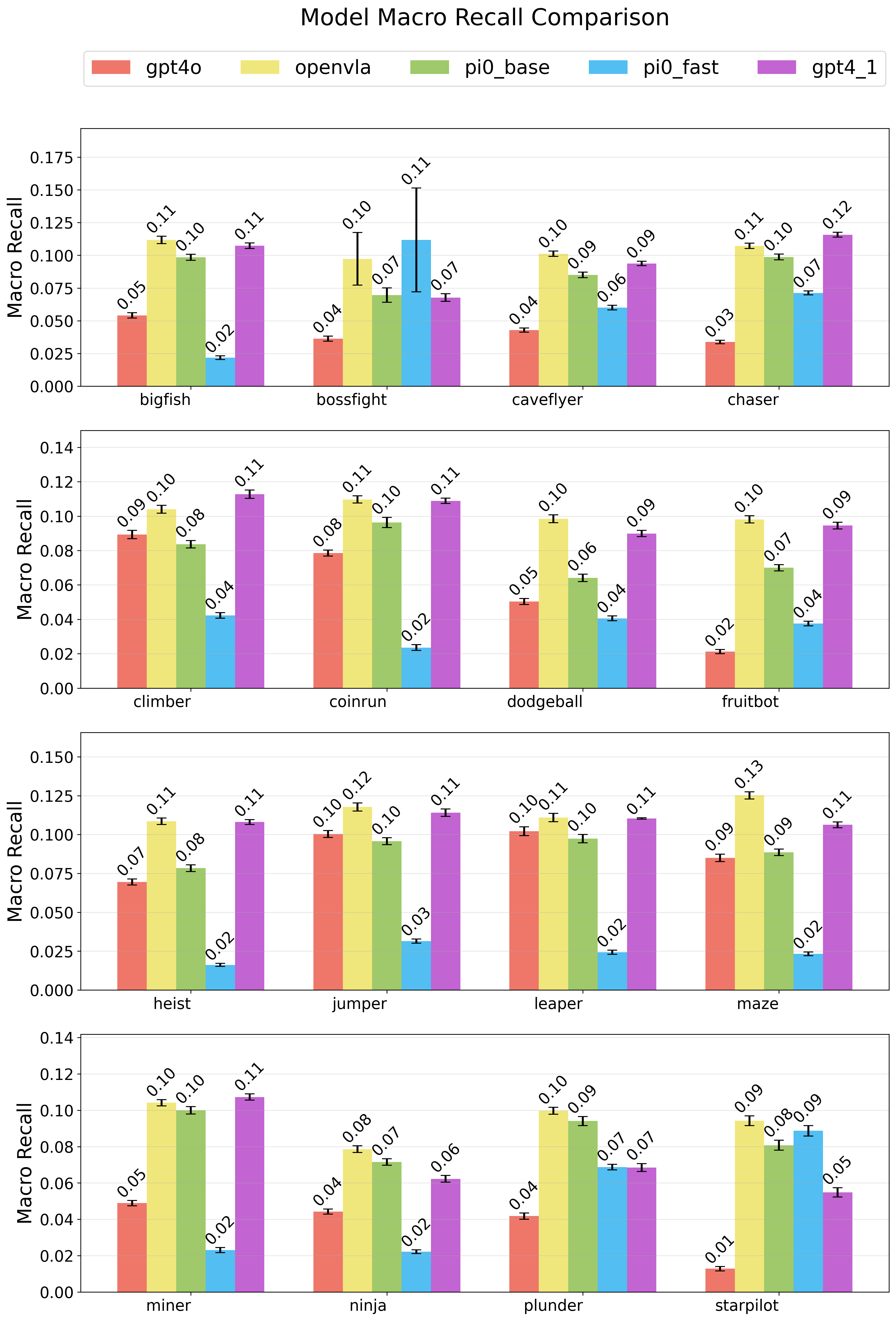}
	\caption{Macro recall across all 5 models. OpenVLA performs better when considering macro recall compared to macro precision, indicating a high number of false positives. GPT 4o shows lower macro recall than precision, indicating biased performance towards specific minority classes. GPT 4.1 and Pi0 Base show relatively less biased and moderate performance, whereas Pi0 FAST showed consistent low recall.}
	\label{fig:fig5}
\end{figure}

Macro recall, as seen in Figure \ref{fig:fig5}, highlighted OpenVLA's aggressive prediction strategy, resulting in consistently high recall but lower precision due to numerous false positives. GPT 4o, despite higher macro precision, showed substantially lower recall, indicating pronounced class biases towards a few classes, and a potentially overly conservative nature about predicting certain classes compared to other classes. GPT 4.1 exhibited a relatively balanced approach, achieving both recall and precision, thus indicating less bias towards specific classes. Pi0 Base displayed a moderately balanced performance without leading in any subdataset, whereas Pi0 FAST consistently showed low recall, reflecting challenges in identifying true positives across all classes.

\begin{figure}
    \centering
    \includegraphics[width=1\linewidth]{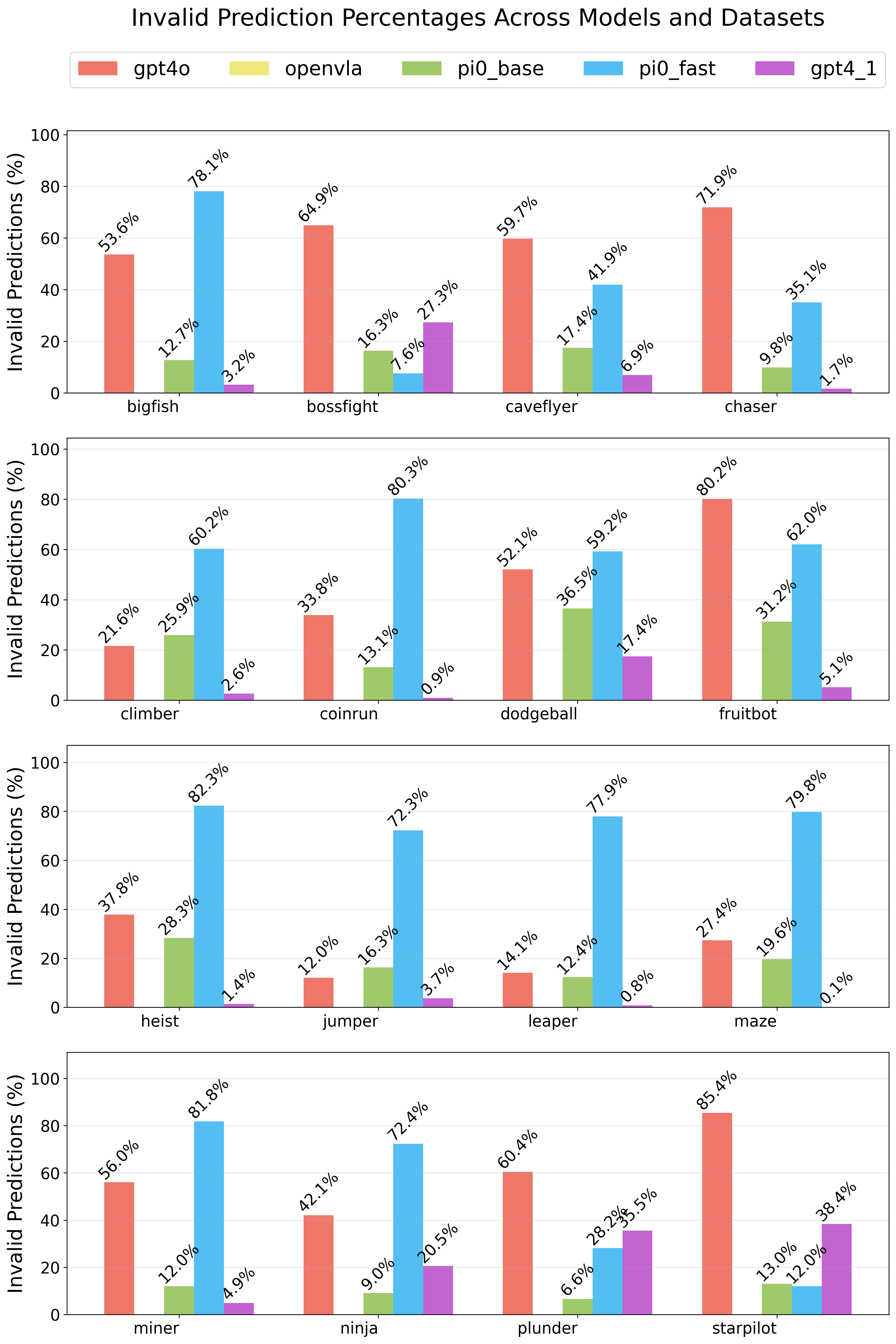}
\caption{Percentage invalids across all 5 models. Invalids refer to model predictions that are not valid actions in the subdataset's action space. Pi0 FAST and GPT 4o struggle to produce valid actions irrespective of the subdataset.}
	\label{fig:fig6}
\end{figure}

The analysis of invalid predictions highlighted significant disparities. OpenVLA, by design, produced zero invalid predictions, ensuring actions always fell within the valid action space. As seen in Figure \ref{fig:fig6}, GPT 4o and Pi0 FAST frequently generated invalid outputs, with some datasets reaching invalid prediction rates above 80\%. 

\subsubsection{Outlier Influence and Majority Prediction Performance}

\begin{figure}
	    \centering
	    \includegraphics[width=9cm]{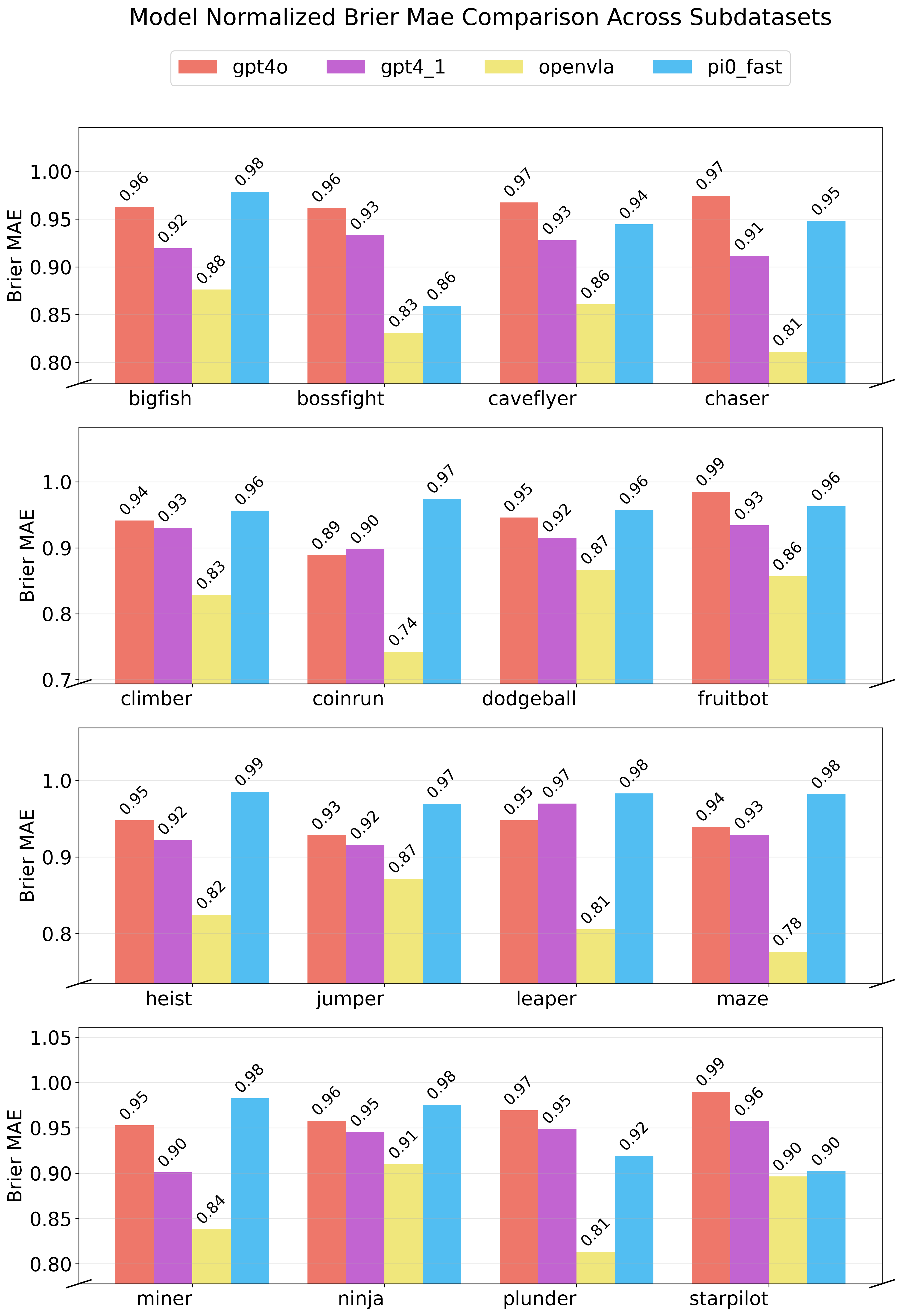}
	    \caption{Normalized Brier MAE across 4 models - GPT 4o, OpenVLA, GPT 4.1, Pi0 FAST. It follows the trends of regular Brier MAE, indicating minimal reliance on outliers.}
	    \label{fig:fig7}
	\end{figure}

\begin{figure}
    \centering
    \includegraphics[width=9cm]{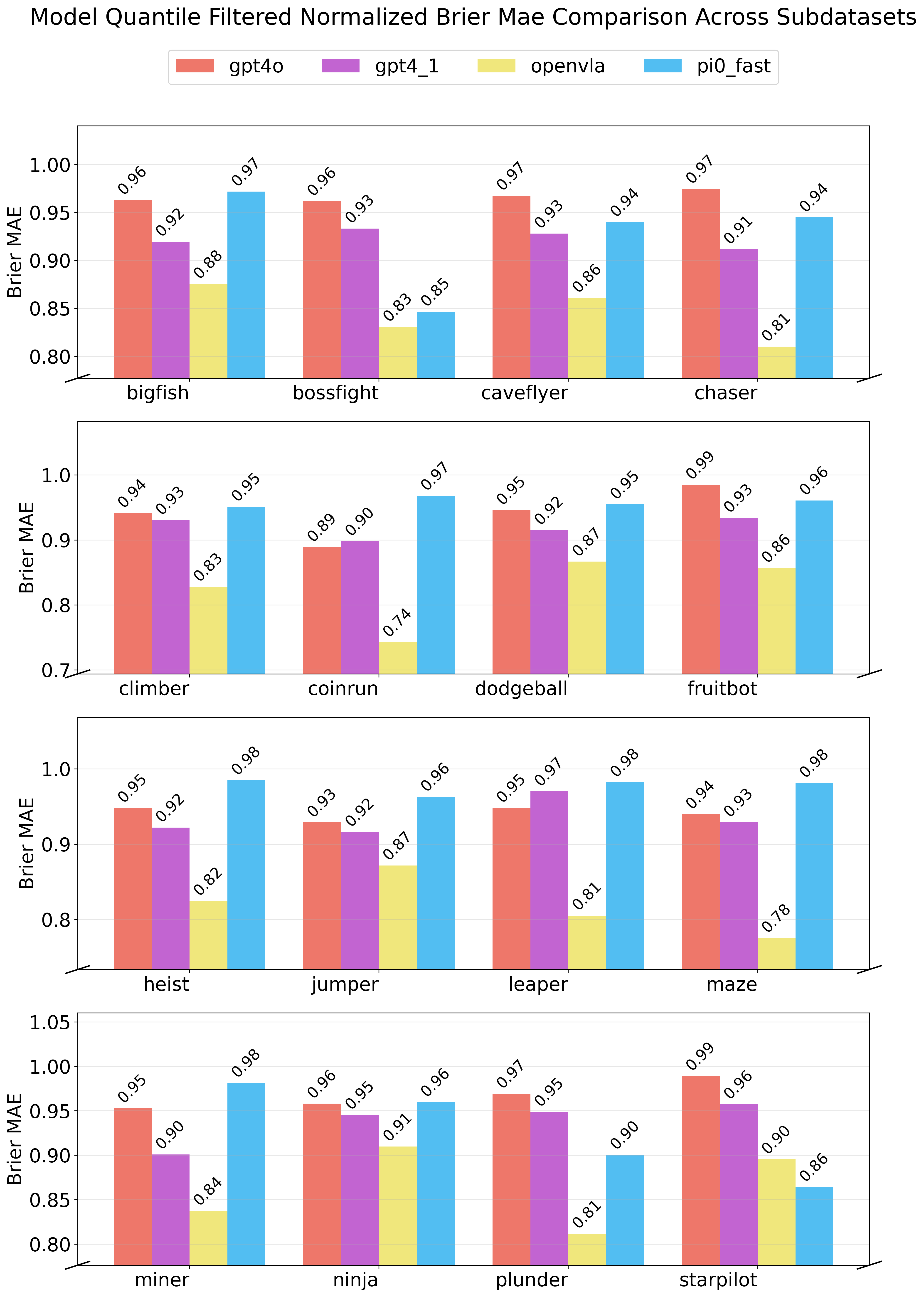}
    \caption{Normalized Quantile Filtered Brier MAE across 4 models. It follows trends of regular and Normalized Brier MAE scores, indicating minimal reliance on extreme predictions.}
    \label{fig:fig8}
\end{figure}

As seen in Figures \ref{fig:fig7} and \ref{fig:fig8}, the normalized Brier MAE and quantile-filtered analysis indicated consistent error trends when compared to regular Brier MAE scores across models and datasets, demonstrating minimal reliance on outlier predictions. Given the already high median errors, the influence of outliers was negligible.

\begin{figure}
    \centering
    \includegraphics[width=1\linewidth]{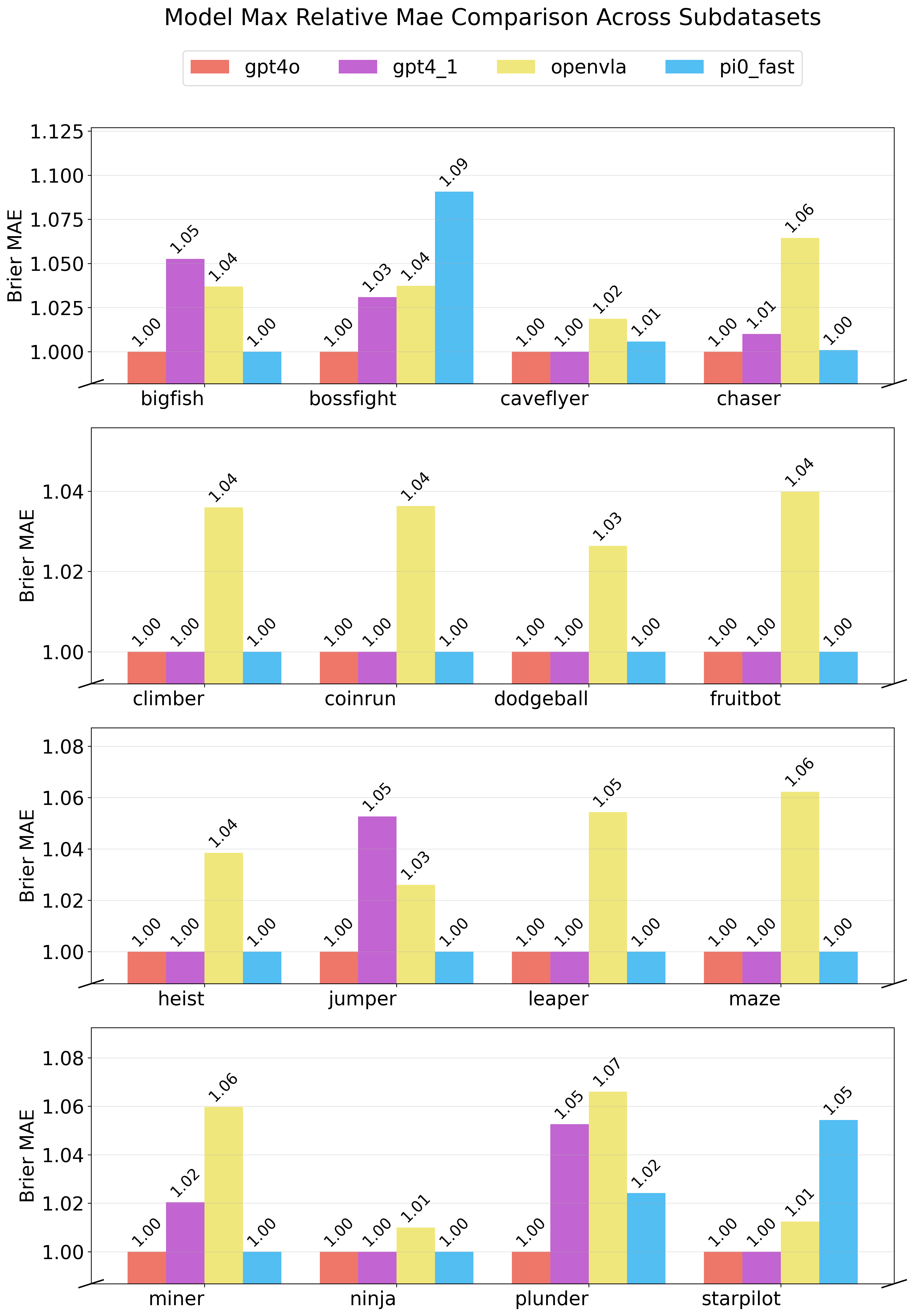}
    \caption{Maximum relative Brier MAE across 4 models. All values are equal to or around 1.0, indicating limited deviation between median and extreme bad predictions.}
    \label{fig:fig9}
\end{figure}

Further investigation using the max relative Brier MAE metric showed minimal variance in model performance, with the scores staying consistently close to 1.0. This finding, seen in Figure \ref{fig:fig9}, suggests limited deviation between median and extreme error values, primarily due to the universally high error levels across all models reducing the relative significance of outlier predictions.

\subsection{Models Struggle with Unseen Discrete Action Data}

\subsubsection{Dataset-Specific Performance Across Models} 

Our evaluation revealed distinct performance patterns across the five models—OpenVLA, GPT-4o, GPT-4.1, Pi0 Base, and Pi0 FAST—when exposed to out-of-distribution (OOD) datasets with discrete action spaces from Procgen.


\begin{figure}
    \centering
    \includegraphics[width=1\linewidth]{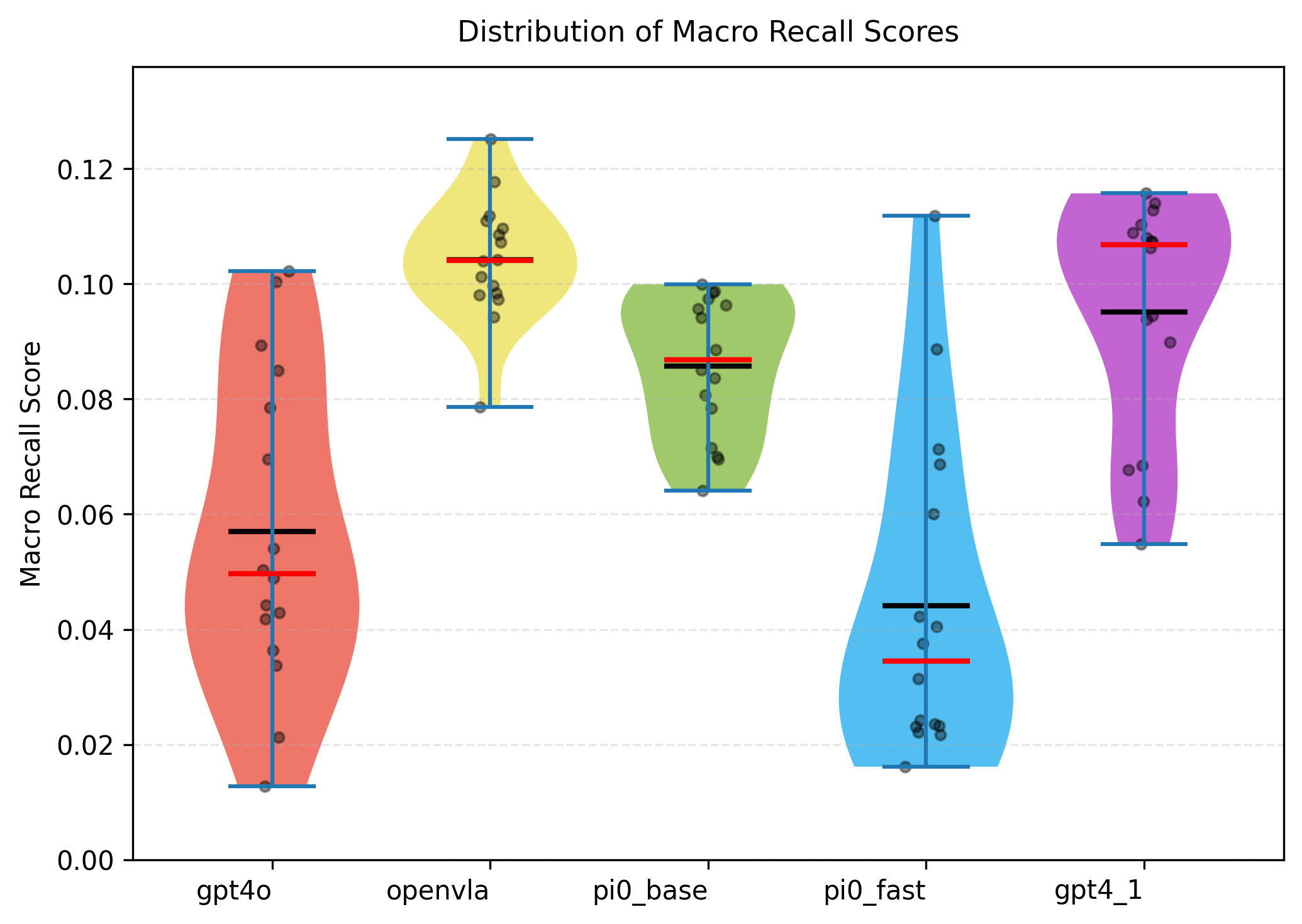}
    \caption{Distribution of model macro recall across subdatasets. The black and red lines indicate the mean and median, respectively.}
    \label{fig:macro_recall_violin}
\end{figure}

\textbf{OpenVLA} consistently achieved the highest macro-recall among the evaluated models across nearly all 16 Procgen subdatasets, typically ranging from 9\% to 12\% as shown in \ref{fig:fig5} and \ref{fig:macro_recall_violin}. It performed notably better on simpler tasks, achieving a maximum recall of approximately 12.5\% on the Maze dataset, characterized by straightforward navigation and collection actions without the complexity of additional special actions. Conversely, OpenVLA's recall dropped to its lowest (~8 percent) on the Ninja dataset, which involves multiple directional special ``fire'' actions, indicating difficulty handling tasks requiring timely execution of specialized actions. This could also indicate increased unfamiliarity with the action type since movement actions may be more similar to the model's training environment than special actions.

\textbf{GPT-4 Series} occupied a middle performance tier, displaying macro-recall values roughly between 6\% and 11\% as seen in \ref{fig:fig10} and \ref{fig:macro_recall_violin}. GPT-4.1 generally outperformed GPT-4o across most datasets. Notably, GPT-4.1 showed superior performance when compared to Pi0 Base on simpler datasets like Leaper, Climber, Jumper, Maze, and Heist. Specifically, GPT-4.1 achieved macro-recall scores of around 11\% on simpler datasets like Leaper, Climber, Jumper, Bigfish, Chaser, Coinrun, Miner, Maze, and Heist. Its performance diminished to around 7\% on more challenging environments such as Plunder, Starpilot, Bossfight, and Ninja.

Conversely, GPT-4o exhibited significantly weaker performance than Pi0 Base on datasets involving timely and sparse execution of special actions. This was particularly evident in datasets involving ``fire'' actions, such as Plunder, Fruitbot, and Starpilot, indicating GPT-4o’s difficulty managing tasks requiring precise timing.  While GPT-4o showed the highest recalls around 8.5\% on datasets with simpler movement-only actions (Leaper, Climber, Jumper, Coinrun, Maze, Heist), it performed significantly worse on more complex tasks. Recall dropped to around 3\% on more complex tasks like Chaser and Bossfight, and below 2.5\% on Fruitbot and Starpilot, both of which feature special actions.

\textbf{Pi0 Base} achieved macro recalls typically between 6\% and 10\%, as seen in \ref{fig:fig10} and \ref{fig:macro_recall_violin}. Its best performance (around 10 percent) appeared on relatively simpler datasets such as Leaper, Plunder, Jumper, Bigfish, Chaser, Coinrun, and Miner. Conversely, macro recall dropped to approximately 6\% on datasets involving special actions like Dodgeball, Fruitbot, Bossfight, and Ninja.

\textbf{Pi0 FAST} exhibited the lowest macro recall among the evaluated models, frequently below 6 percent, and occasionally as low as 2\% on datasets including Leaper, Bigfish, Coinrun, Ninja, Miner, Maze, and Heist as seen in \ref{fig:fig10} and \ref{fig:macro_recall_violin}. However, Pi0 FAST displayed anomalously higher recalls (~10 percent) on datasets such as Starpilot and Bossfight. Despite these datasets involving special actions, the actions are executed frequently rather than sparsely, aligning better with Pi0 FAST's predictions and ground truth distributions.

In general, our findings highlight a clear difficulty across models in accurately handling datasets that require complex, sparse, or timely executed actions, particularly when special non-movement actions are involved. The macro recall performance appears directly correlated to how closely the distribution of model predictions aligns with the distribution of the ground truth actions.

\subsubsection{Action-Class Specific Performance Across Models}
\begin{figure}
    \centering
    \includegraphics[width=\linewidth]{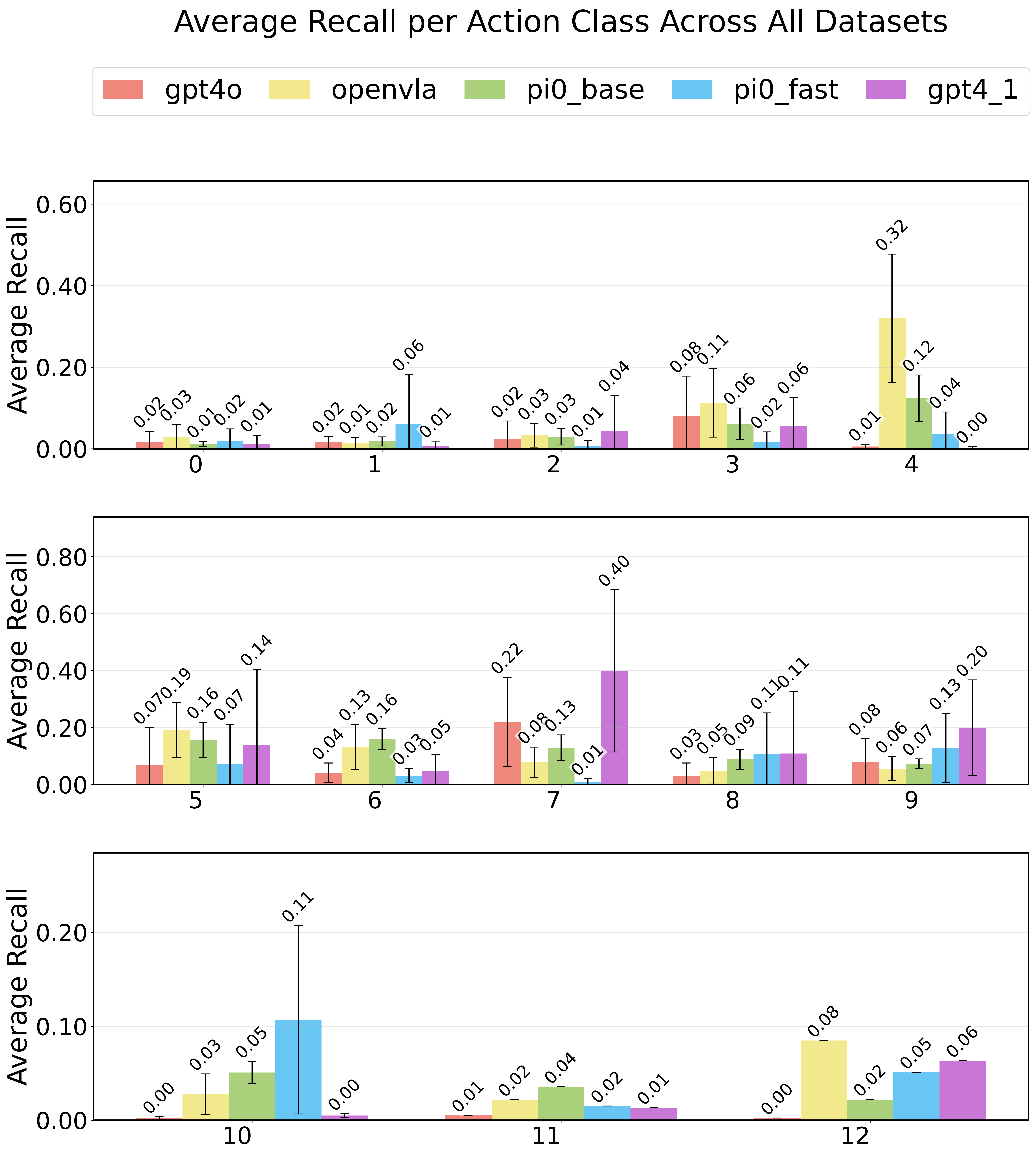}
    \caption{Class-wise recall averaged across all subdatasets for all 5 models. Recall and Variance on classes 4-10 is relatively higher, and is low on classes 0-3, and 11-12.}
    \label{fig:fig11}
\end{figure}

A detailed analysis of recall performance by individual action classes offers valuable insights into model-specific strengths, weaknesses, and intrinsic biases in predicting discrete actions. As seen in Figure \ref{fig:fig11}, action classes 4 through 10 generally displayed higher recall scores and greater variance among models, indicating varying degrees of predictive confidence. In contrast, action classes 0 to 3 and 11 to 12 consistently yielded low recall values across all models, demonstrating limited predictive capability. 
\begin{figure}
    \centering
    \includegraphics[width=1\linewidth]{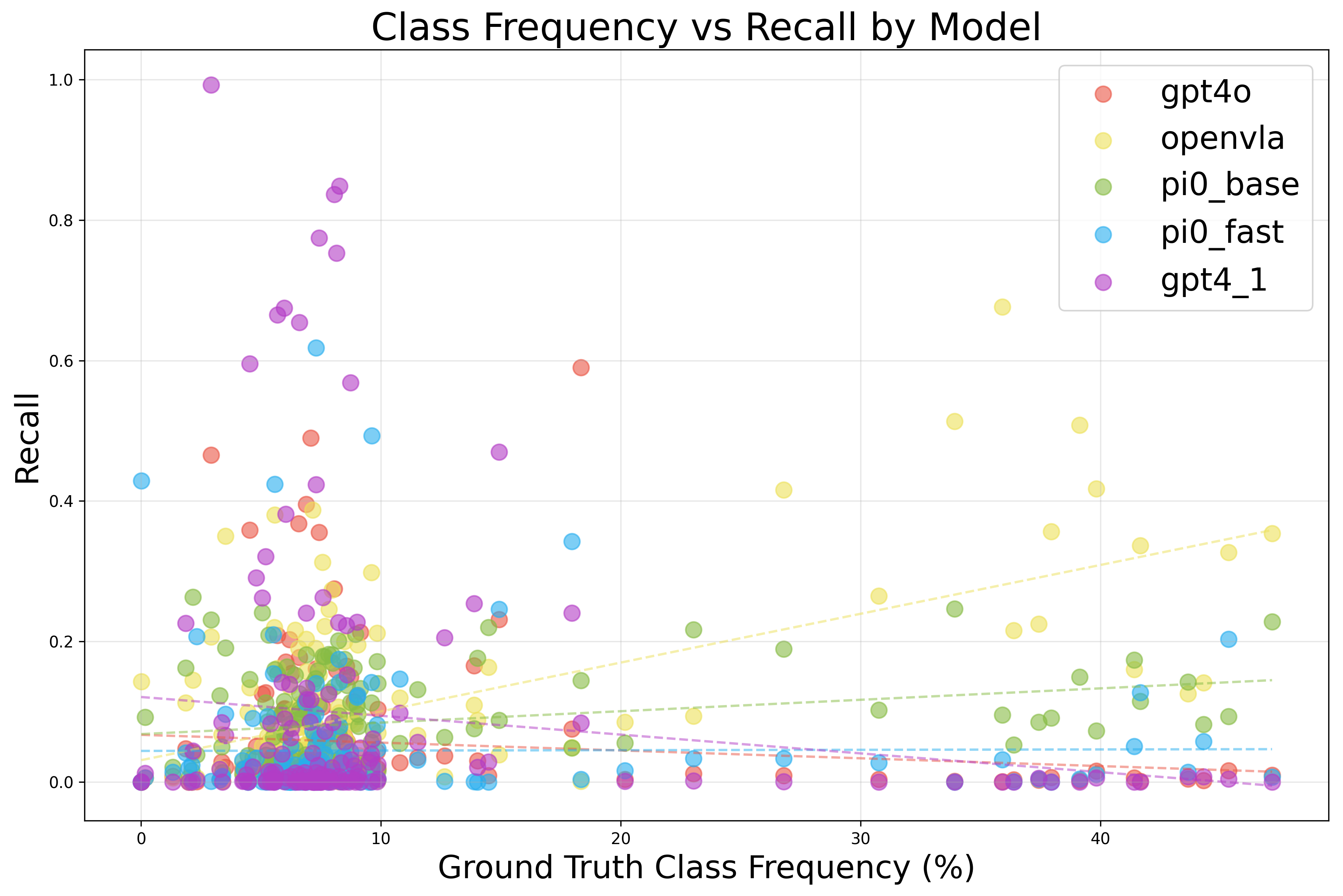}
    \caption{Model action class recall vs. action class frequency} 
    \label{fig:recall_vs_class_frequency}
\end{figure}
When comparing the model action class recall against action class frequency in Figure \ref{fig:recall_vs_class_frequency}, OpenVLA exhibits the strongest positive correlation, attaining the highest recall on the most prevalent classes, while Pi0 base shows little correlation. Both GPT-4o and GPT-4.1 display slightly negative trends, indicating more struggle with common classes than with rarer ones. Pi0 Fast's recall remains largely constant across action class frequencies, indicating uniform yet low performance.

\begin{figure}
    \centering
    \includegraphics[width=1\linewidth]{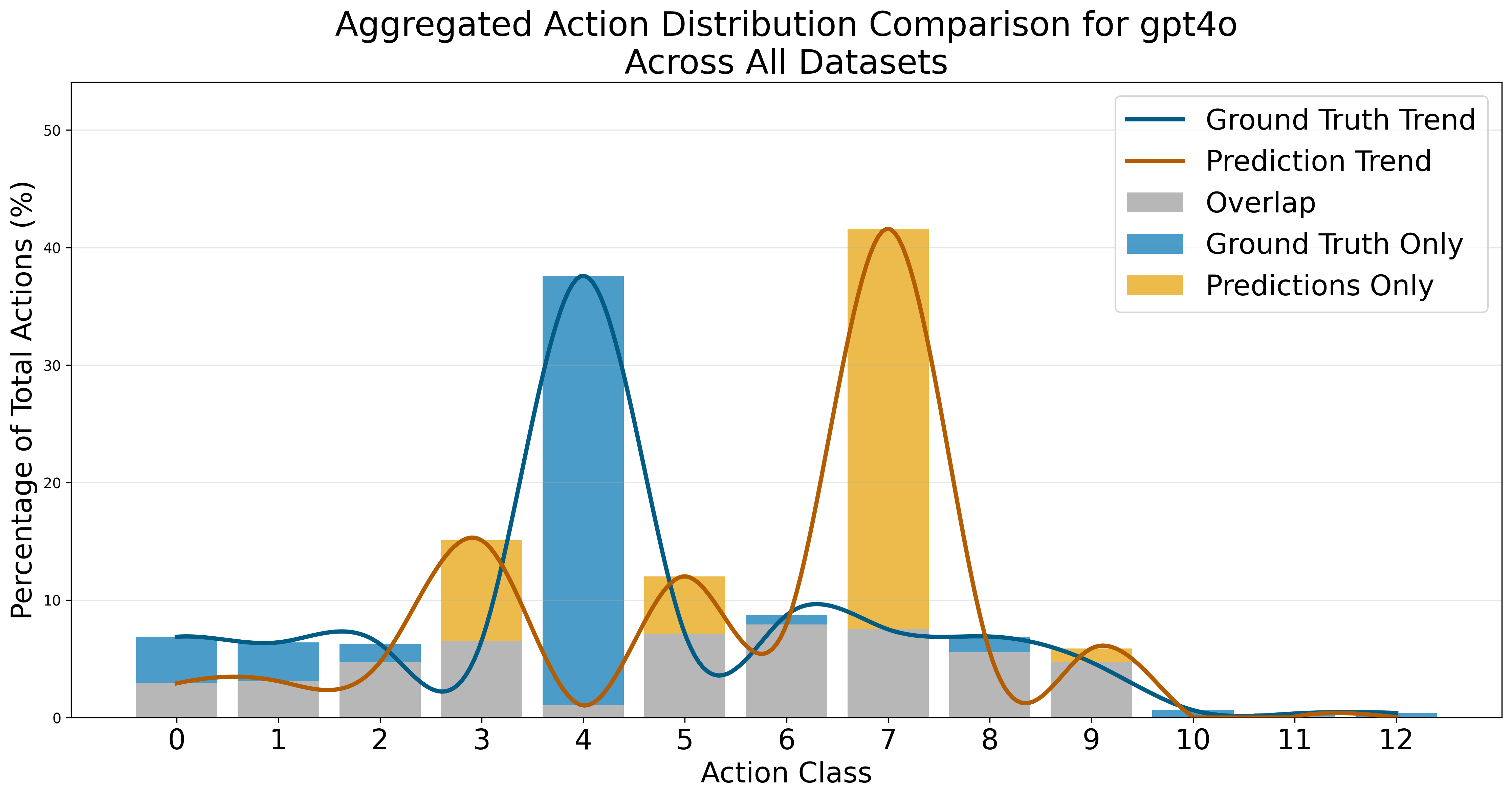}
    \caption{GPT4o prediction vs. ground truth distributions. }
    \label{fig:fig12}
\end{figure}

GPT-4o exhibited notable recall performance on action class 7(``Right'' movement), achieving a high recall of approximately 0.22 as seen in \ref{fig:fig11}. This suggests a directional bias likely influenced by the prevalent rightward progression in many Procgen tasks. Conversely, GPT-4o struggled significantly with actions 4, 10, 11, and 12, displaying near-zero recall, highlighting clear limitations in handling certain peripheral or specialized actions. Specifically, the difference between the most frequent action classes in predicted actions and ground truth actions can be seen in Figure \ref{fig:fig12}.

\begin{figure}
    \centering
    \includegraphics[width=1\linewidth]{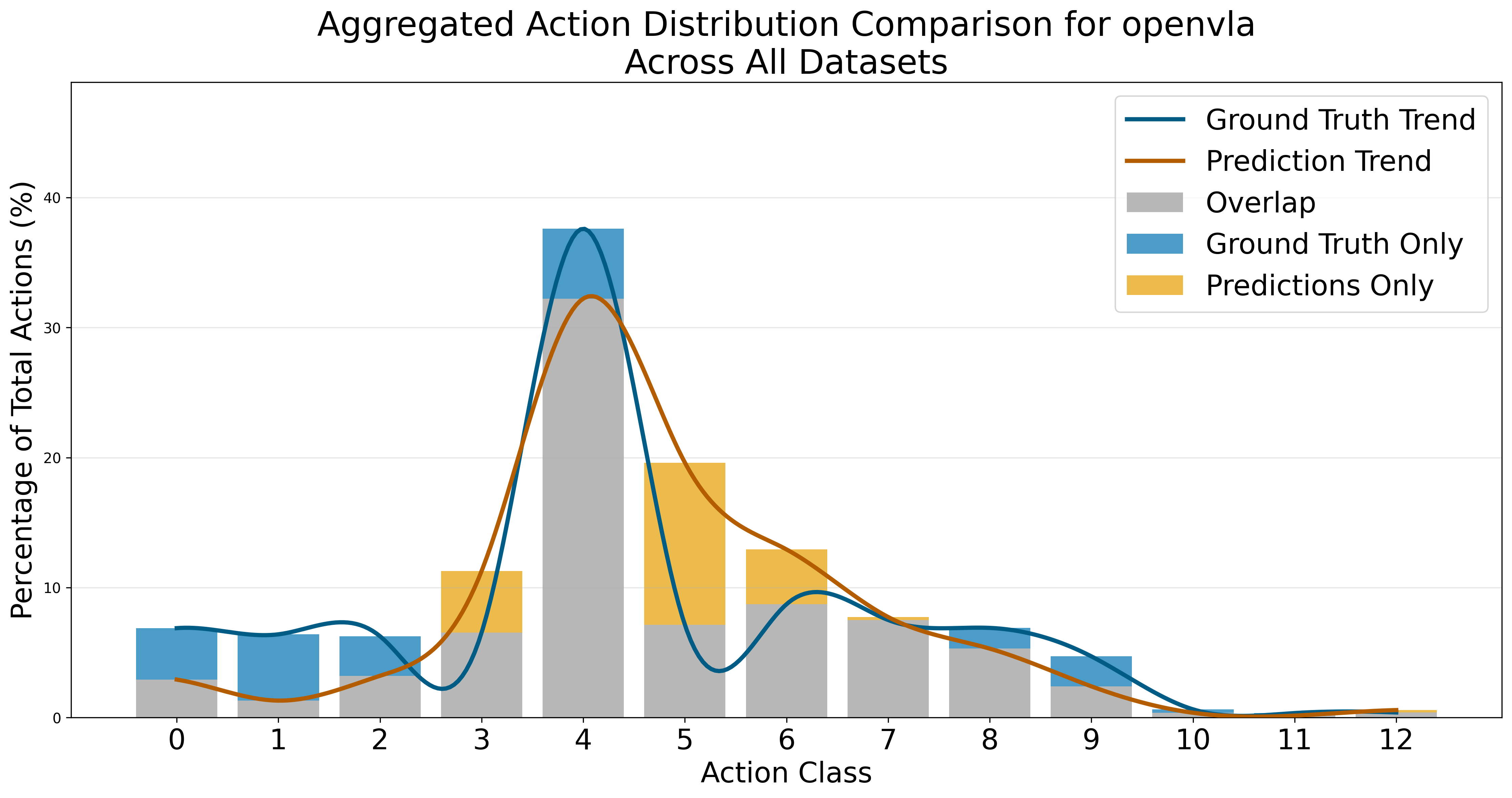}
    \caption{OpenVLA prediction vs. ground truth distributions}
    \label{fig:fig13}
\end{figure}

OpenVLA predominantly favored action class 4 (``Do Nothing'') as seen in \ref{fig:fig13}, recording the highest recall (~0.32) as seen in \ref{fig:fig11}. This bias arises from a combined influence of its decoding mechanism and Procgen action space sizes. Most Procgen datasets have an action space of size 9 or 10, and due to unnormalization of OpenVLA’s predicted actions using the dataset statistics, 4 can correspond to 0 in the normalized range, which is the center of OpenVLA's output distribution. OpenVLA displays considerable difficulty predicting peripheral classes (0, 1, 8, 10, and 11), with recall scores as low as 0.03, indicating a substantial predictive bias toward central actions.

\begin{figure}
    \centering
    \includegraphics[width=1\linewidth]{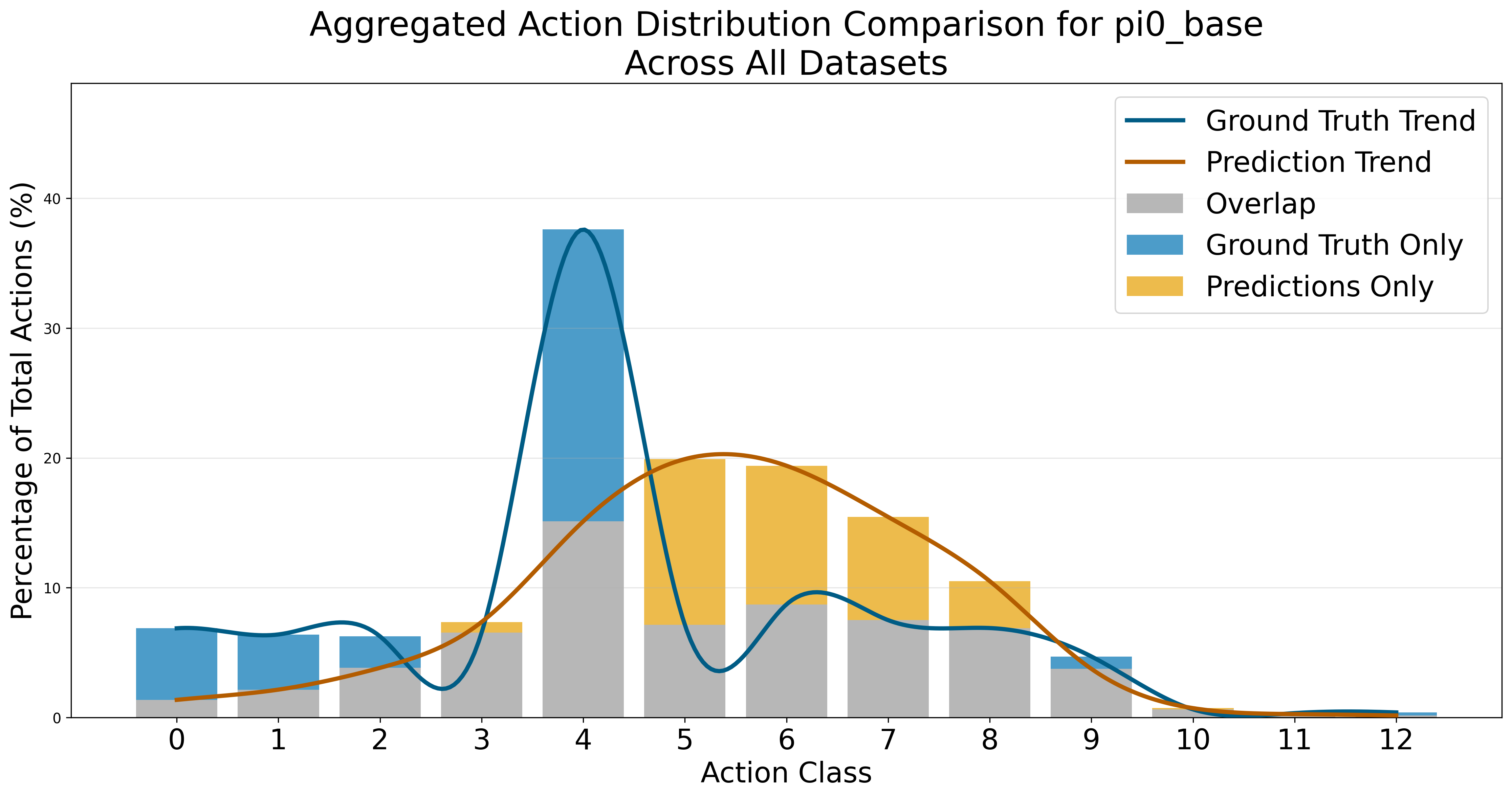}
    \caption{Pi0 base prediction vs. ground truth distributions}
    \label{fig:fig14}
\end{figure}

Pi0 Base similarly gravitated toward central action classes (4, 5, 6, and 7) as seen in \ref{fig:fig14} due to its normalization-unormalization decoding approach, resulting in moderate recall values (~ 0.12 - 0.16) as seen in \ref{fig:fig11}. Its recall sharply declined for peripheral and special action classes (0, 1, 2, and 12), dropping to approximately 0.02. Compared to OpenVLA, Pi0 Base’s diffusion-based approach led to less concentrated predictions, distributing recall more evenly but still inadequately across peripheral classes.

\begin{figure}
    \centering
    \includegraphics[width=1\linewidth]{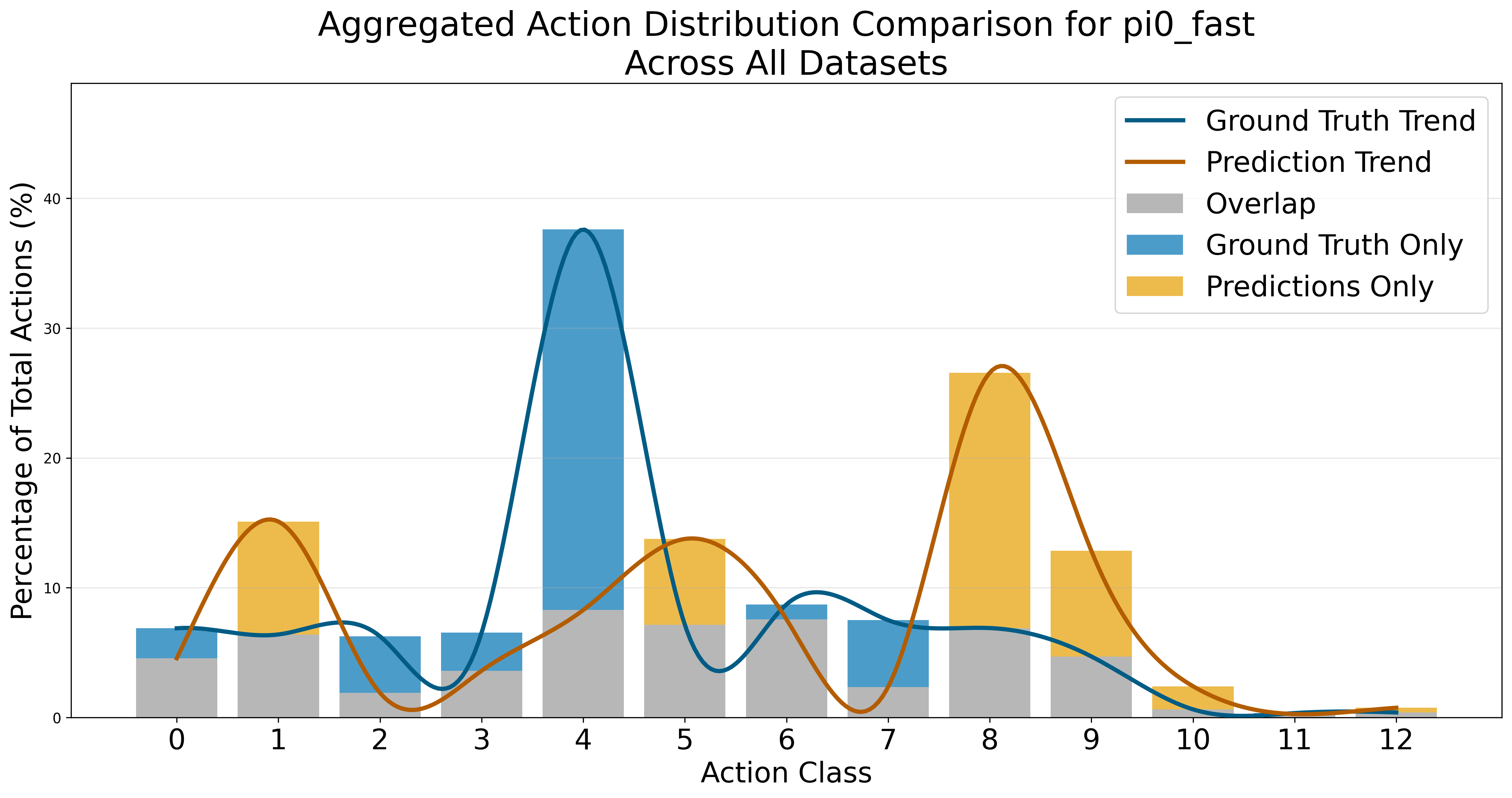}
    \caption{Pi0 fast prediction vs. ground truth distributions}
    \label{fig:fig15}
\end{figure}

Pi0 FAST displayed a pronounced bias towards action classes (1, 5, 8, and 9) as seen in \ref{fig:fig15}, with action percentage between 10\% and 26\%. This bias reflects its difficulty generalizing to OOD data, often producing higher normalized action predictions. Conversely, Pi0 FAST severely underperformed on actions between the above prevalent classes, such as 0, 2, 3, 7, and 11, frequently achieving near-zero recall as seen in \ref{fig:fig11}, emphasizing its limitations in balanced action class predictions.

\begin{figure}[H]
    \centering
    \includegraphics[width=1\linewidth]{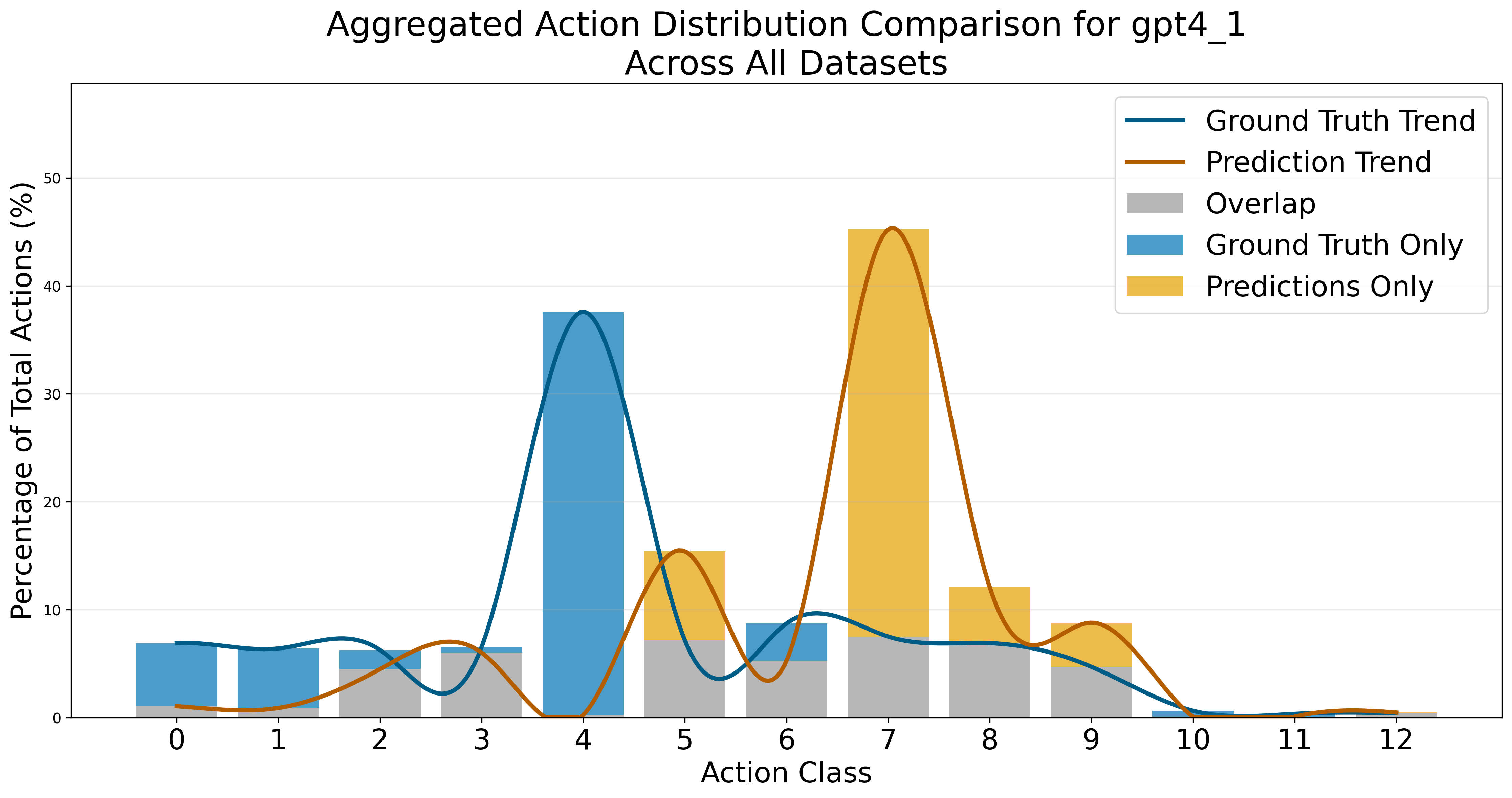}
    \caption{GPT-4.1 prediction vs. ground truth distributions}
    \label{fig:fig16}
\end{figure}

GPT-4.1 demonstrated a distinct bias toward action class 7 (``Right'' movement) as seen in \ref{fig:fig16}, achieving exceptionally high recall (~0.40) as seen in \ref{fig:fig11}, consistent with the directional tendencies of many Procgen environments. It also showed moderate recall (~0.20) for special action class 9. However, GPT-4.1 struggled notably with several other action classes, particularly actions 0, 1, 4, 10, and 11, consistently resulting in minimal recall (~0.01).

These results clearly indicate model-specific biases towards particular action classes, underscoring the importance of targeted adjustments in model training and decoding methods to foster more balanced and generalized performance across the entire action space.

\FloatBarrier
\subsubsection{Models' Prediction Collapse to Few Action Classes}

A consistent observation across all evaluated models was their tendency to collapse predictions predominantly into a limited subset of action classes. Analysis of union confusion matrices as seen in Figure \ref{fig:fig17} and action distributions \ref{fig:fig12}, \ref{fig:fig13}, \ref{fig:fig14}, \ref{fig:fig15}, and \ref{fig:fig16}  aggregated over all 16 Procgen subdatasets highlighted clear default prediction behaviors.

Specifically, each model exhibited distinct biases toward particular default actions:

\begin{itemize}
  \item OpenVLA frequently defaulted to predicting action classes 4 (``Do Nothing''), 5 (``UP''), and 6 (``RIGHT + DOWN'').
  \item Pi0 Base predominantly predicted action classes clustered between 4 to 7 (``Do Nothing'', ``UP'', ``RIGHT + DOWN'', ``RIGHT'') and the range shifts to between 5 to 8 (``UP'', ``RIGHT + DOWN'', ``RIGHT'', ``RIGHT + UP'') when the ground truth actions are between 9 and 12.
  \item Pi0 FAST consistently favored action classes 6 (``RIGHT + DOWN'') and 8 (``RIGHT + UP'').
  \item GPT-4o exhibited a strong bias towards action class 7 (``RIGHT'').
  \item GPT-4.1 frequently defaulted to action classes 7 (``RIGHT'') and 2 (``LEFT + UP'').
\end{itemize}

\begin{figure}
  \centering
  \begin{minipage}[t]{0.5\textwidth}
    \includegraphics[width=\textwidth]{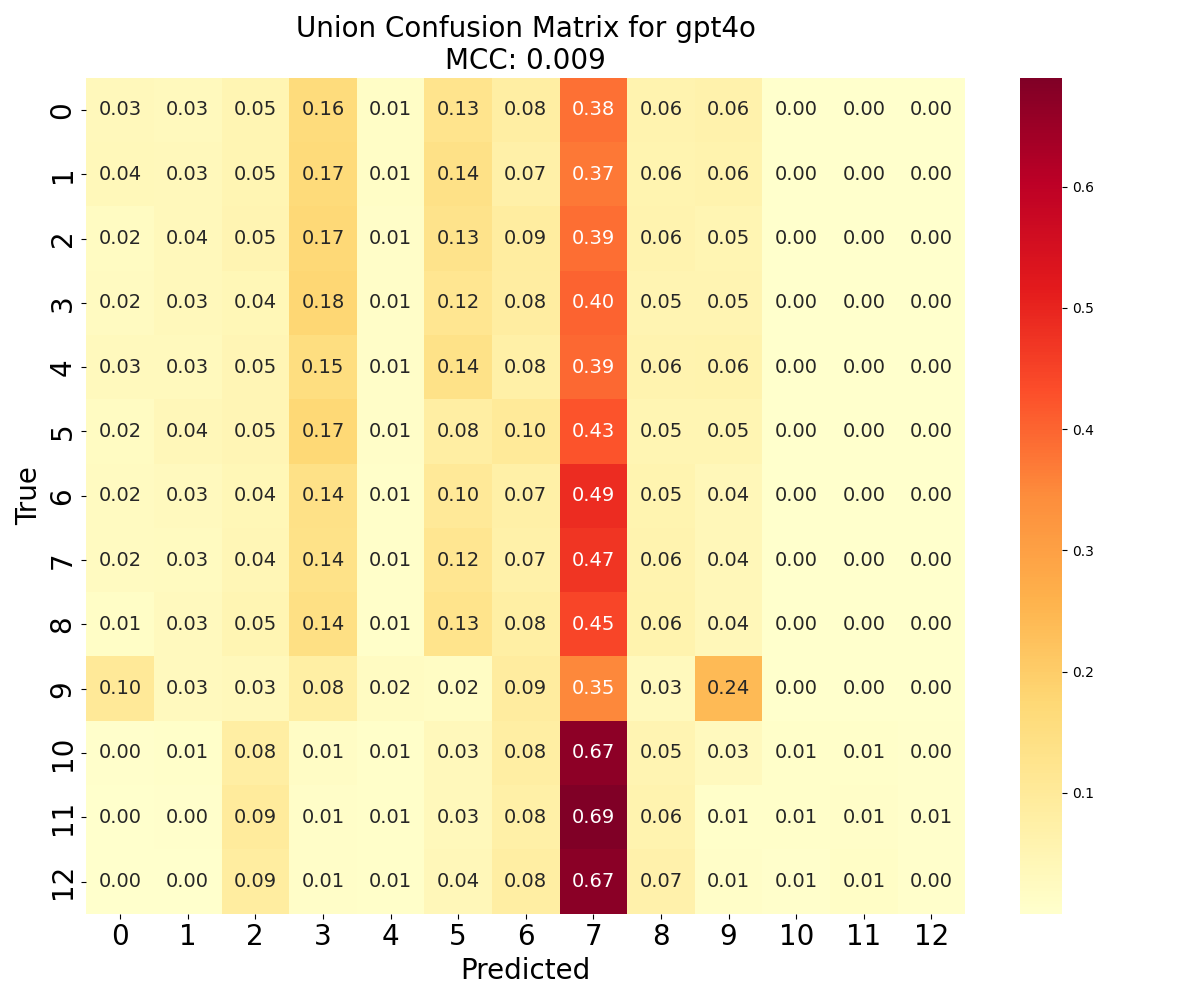}
  \end{minipage}
  \hfill
  \begin{minipage}[t]{0.5\textwidth}
    \includegraphics[width=\textwidth]{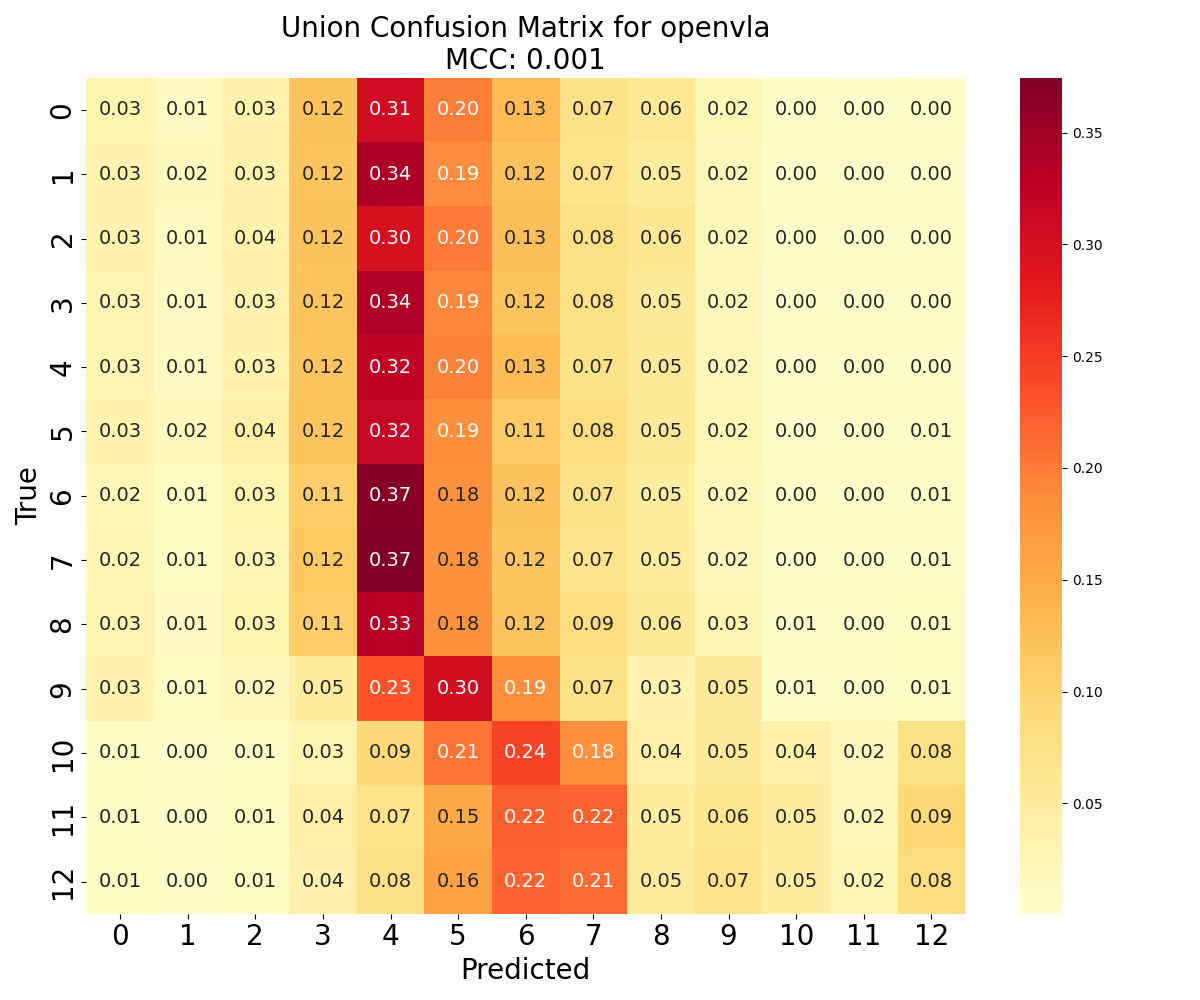}
  \end{minipage}
  \hfill
  \begin{minipage}[t]{0.5\textwidth}
    \includegraphics[width=\textwidth]{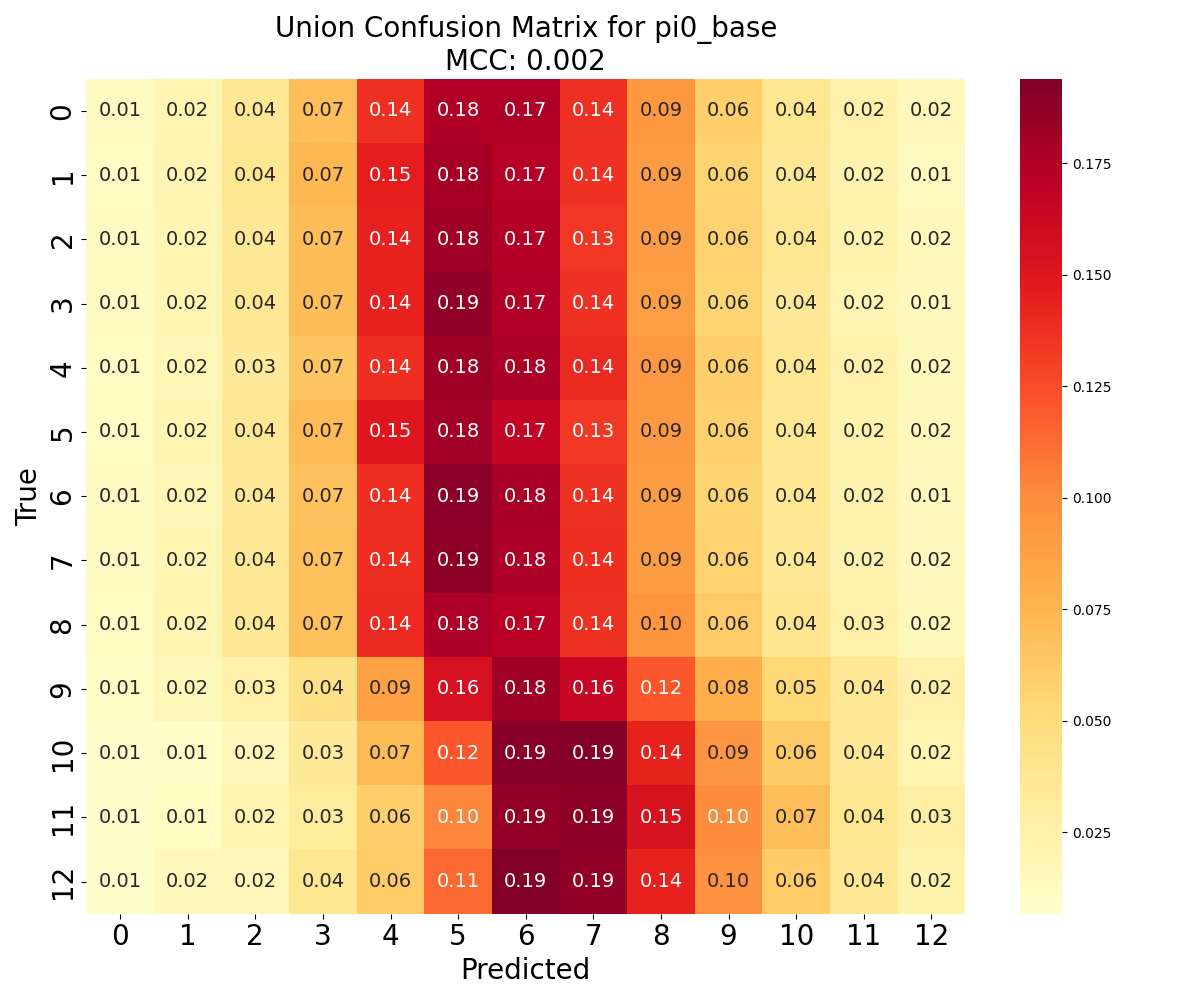}
  \end{minipage}
  \end{figure}
  \begin{figure}
  \begin{minipage}[t]{0.5\textwidth}
    \includegraphics[width=\textwidth]{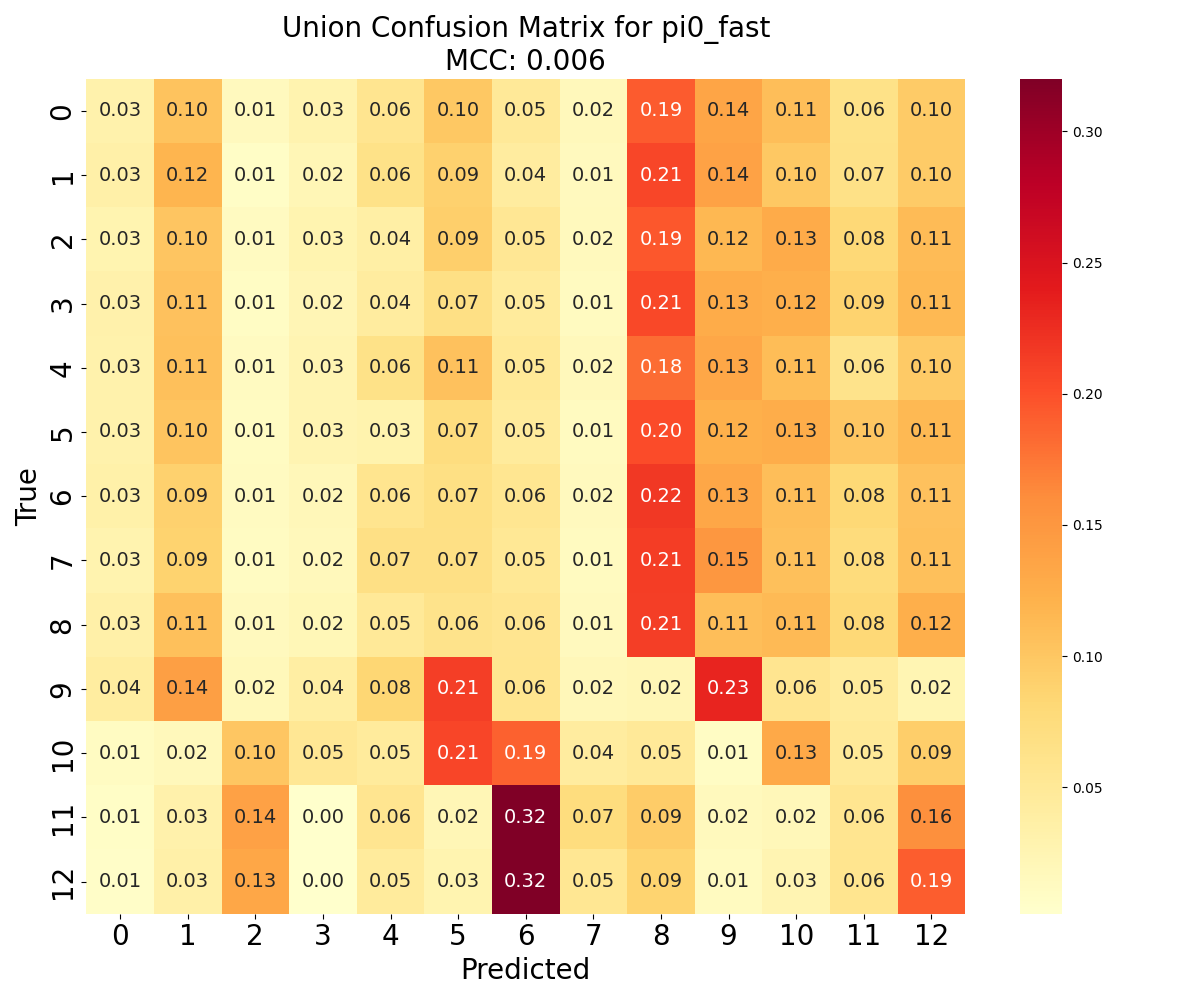}
  \end{minipage}
  \hfill
  \begin{minipage}[t]{0.5\textwidth}
    \includegraphics[width=\textwidth]{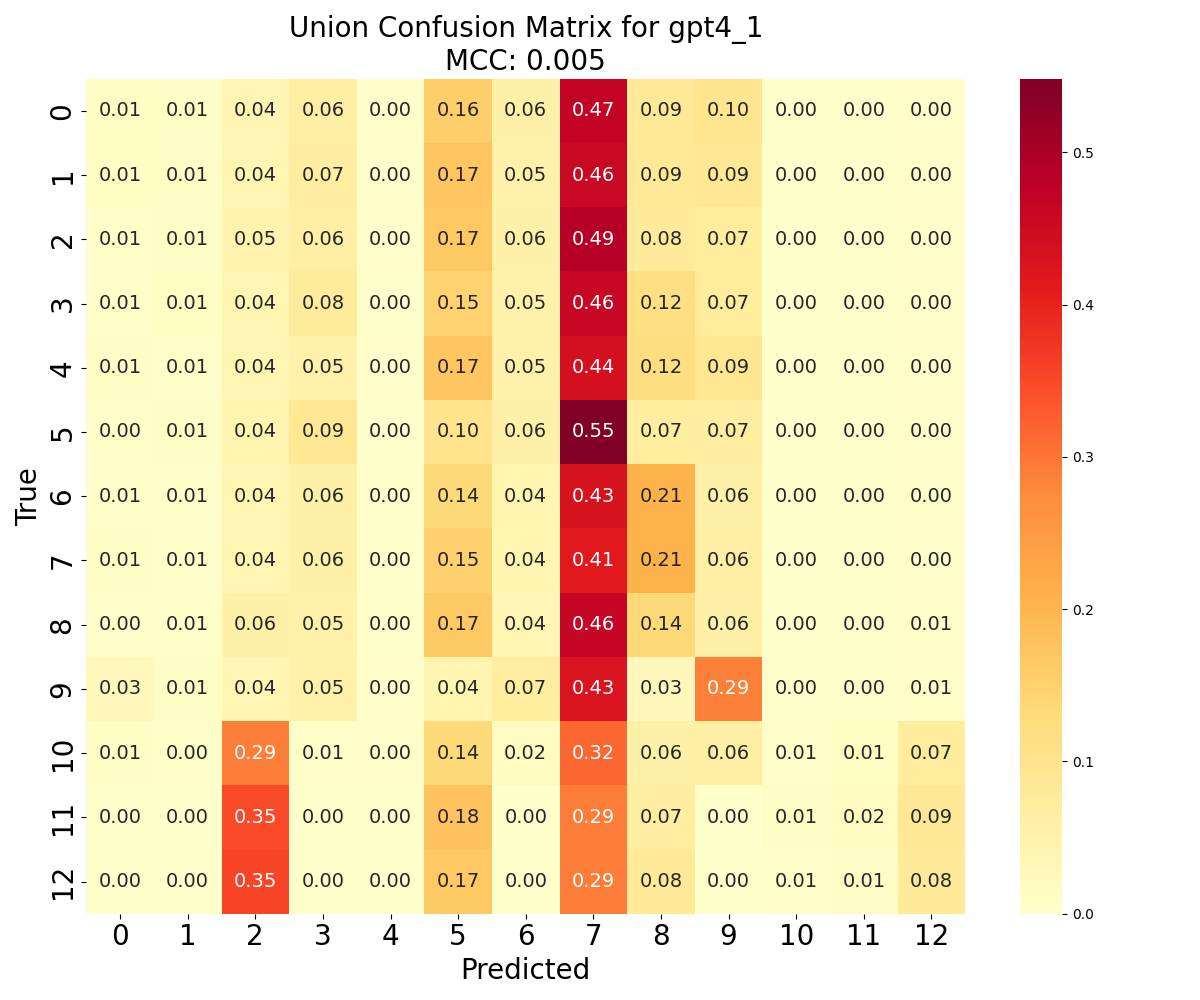}
  \end{minipage}
  \hfill
  \caption{Union confusion matrices depicting frequency of class predictions and ground truth matches. Order from top to bottom - GPT 4o, OpenVLA, Pi0 Base, Pi0 FAST, GPT 4.1. }
\label{fig:fig17}
\end{figure}

\begin{figure}
    \centering
    \includegraphics[width=\linewidth]{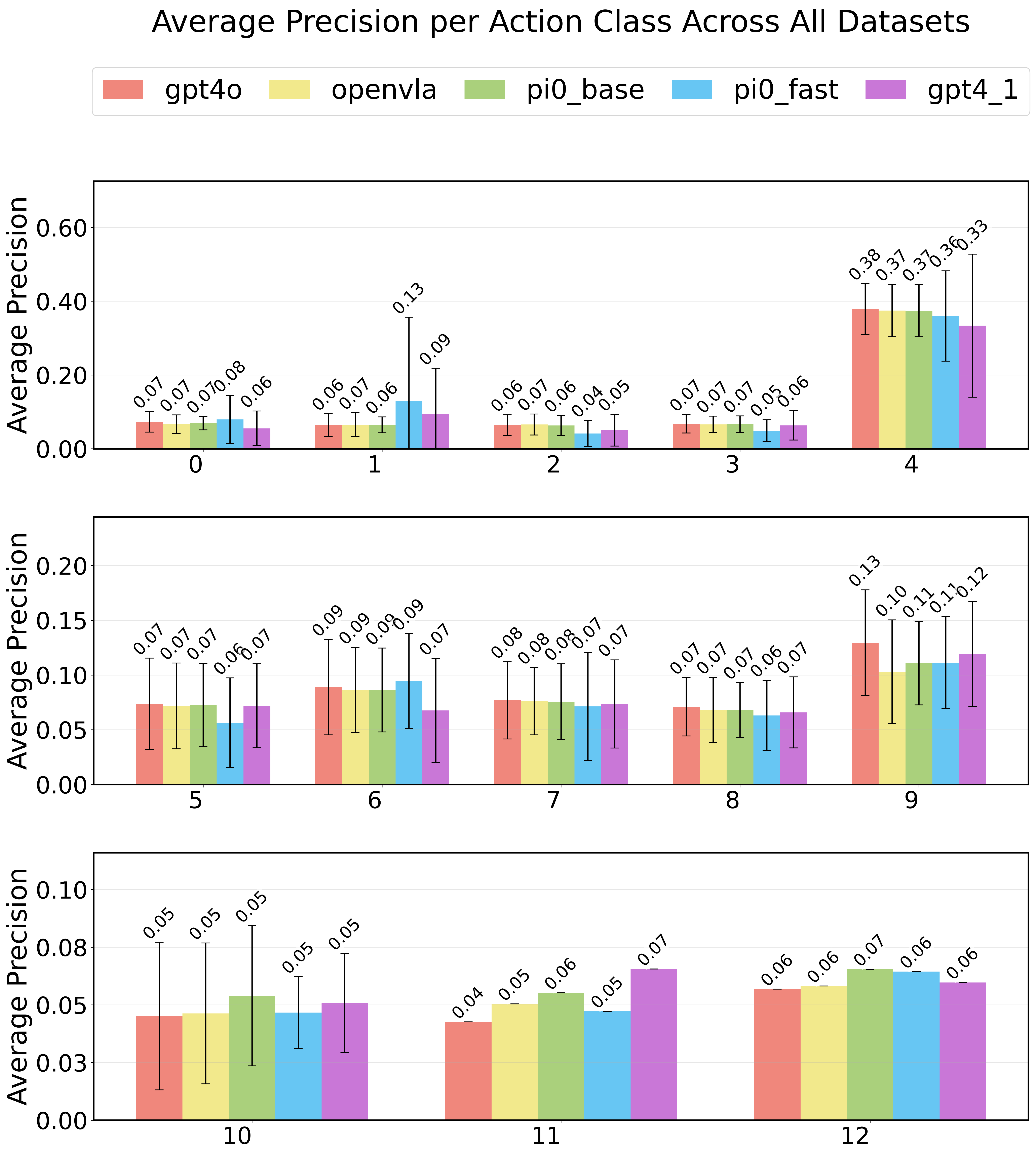}
    \caption{Class-wise precision averaged across all subdatasets for all 5 models. Precision is significantly higher for 4 across all models.}
    \label{fig:fig18}
\end{figure}

Notably, as seen in Figure \ref{fig:fig18}, across all models and subdatasets, action class 4 (``Do Nothing'') exhibited significantly higher precision compared to other classes. We speculate that this elevated precision for class 4 originates from its role as a default assignment for expert actions that fell outside the environment-specific action spaces during the generation of ground truth labels, thus inflating its representation.

For VLMs like GPT-4o and GPT-4.1, the frequent choice of action class 4 may also reflect insufficient understanding of the task, causing it to be uncertain in its predictions and default to ``Do Nothing''. Additionally, the observed strong preference for ``RIGHT'' (action class 7) aligns with the prevalent directional progression in many Procgen environments.

In contrast, OpenVLA, Pi0 Base, and Pi0 FAST primarily defaulted towards actions situated near the center of their normalized prediction range (typically zero). These central predictions commonly map to action class 4 due to rounding effects and the typical size ($<$= 10 actions) of Procgen datasets.

These findings underscore critical biases inherent in model designs and highlight opportunities for refining training approaches and decoding methods to achieve balanced action distributions and improve generalization to unseen action spaces.

\subsection{Models’ Predictions Collapse on Specific Action Classes with OOD Data}

\subsubsection{Patterns of Collapse Influenced by Decoding Techniques}
The behavior of models on out-of-distribution (OOD) data revealed clear patterns of prediction collapse, significantly influenced by their underlying decoding strategies. We specifically examined how diffusion sampling and autoregressive (AR) decoding methods shaped these patterns.

Diffusion models inherently differ from AR models due to their generative approach, which aims to reconstruct data from noise by modeling the entire distribution of training data \cite{ho2020denoisingdiffusionprobabilisticmodels}. In multi-class classification tasks with mutually exclusive single-correct answers, diffusion models respond distinctly when encountering OOD inputs. Primarily, diffusion models excel at reconstructing data similar to their training distribution. However, upon encountering OOD inputs, their reconstruction accuracy deteriorates markedly due to difficulty in mapping the unfamiliar inputs onto the learned data manifold. Consequently, predictions on OOD data are characterized by uncertainty, with predictions distributed across multiple classes instead of sharply peaking at a single class. Visual inspection of the confusion matrices from Pi0 Base on datasets like Bigfish, Climber, and Maze as seen in Figures \ref{fig:fig30}, \ref{fig:fig31}, and \ref{fig:fig32}, supports this finding, demonstrating distributed prediction frequencies rather than dominance by any one class.

In contrast, AR models sequentially predict outputs conditioned on previous outputs, typically applying softmax at the final step for classification tasks. A known limitation of this approach is an inherent tendency toward overconfidence in OOD scenarios. Despite inputs being significantly divergent from training data, AR models frequently produce sharply peaked probability distributions, arbitrarily favoring one class. This behavior arises from their inductive biases, where the softmax activation overestimates probabilities for certain classes. This issue is evidenced clearly in the union confusion matrices seen in Figure \ref{fig:fig17} and the action distributions seen in \ref{fig:fig12}, \ref{fig:fig13}, \ref{fig:fig14}, \ref{fig:fig15}, and \ref{fig:fig16}. Each AR model consistently collapses its predictions onto particular classes, even with nonsensical or significantly divergent inputs, as observed in the individual per-subdataset confusion matrices in the Appendix section \ref{appendix_confusionmatrix} for OpenVLA’s outcomes on Chaser and Jumper, Pi0 FAST’s results on Bigfish, Chaser, and Heist, and GPT 4x’s performance on Heist, Maze, Coinrun and Miner.

\begin{figure}
    \centering
    \includegraphics[width=1\linewidth]{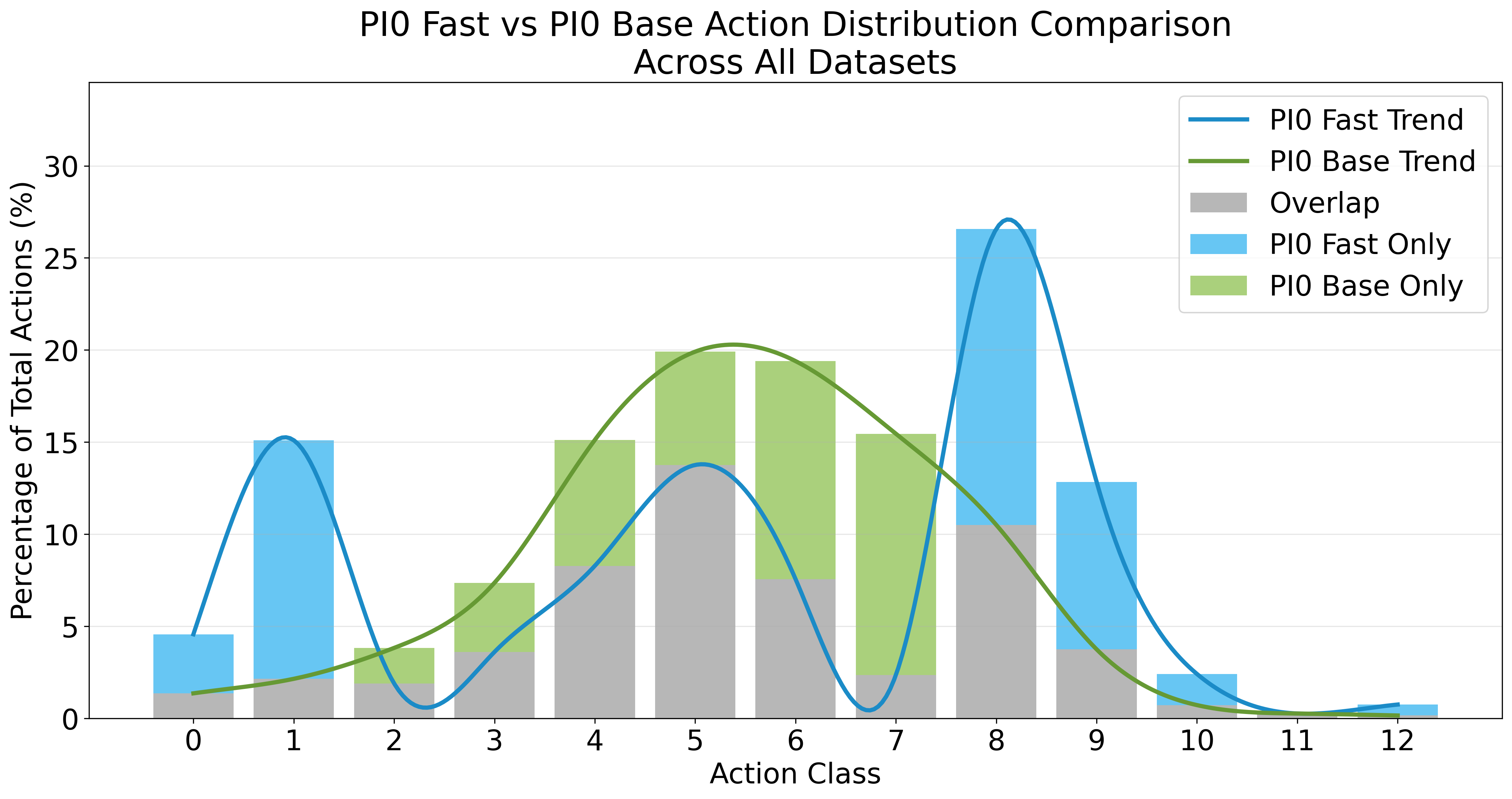}
    \caption{Pi0 Base vs. Pi0 FAST action distributions.  }
    \label{fig:pi0vspi0fast}
\end{figure}

The difference in collapse behaviors between the AR and Diffusion models can be further observed by comparing the action distributions of Pi0 Base and Pi0 FAST in Figure \ref{fig:pi0vspi0fast}, as both models share the PaliGemma VLM backbone in their architecture.
The prediction collapse behavior also varies significantly between different AR models due to differences in tokenization methods. Specifically, OpenVLA and Pi0 FAST differ fundamentally in their tokenization and decoding strategies, influencing their OOD performance distinctly.

Pi0 FAST utilizes the Discrete Cosine Transform (DCT) \cite{dct} in combination with Byte Pair Encoding (BPE) \cite{sennrich2016neuralmachinetranslationrare}, focusing on low-frequency components that represent the overarching shape of the signals. This frequency-domain approach, particularly the strategy of flattening DCT coefficients prioritizing low-frequency information, results in concentrated predictions due to stronger global pattern capture and robust representation against input noise. This approach enhances consistency and stability in action predictions, as evidenced by concentrated class predictions observed in the confusion matrices.

Conversely, OpenVLA employs a discrete binning approach, mapping continuous action spaces to discrete bins directly, thus not capturing global patterns in any sense when evaluated in a zero-shot setting. OpenVLA exhibits sensitivity to variations and input noise, resulting in more evenly distributed probability mass across action classes, as reflected in the broader spread observed in its confusion matrices.

Overall, our analysis highlights how decoding methods and tokenization strategies critically impact models' behaviors in OOD scenarios, indicating the need for targeted adjustments depending on the specific application requirements for generalization performance.

\section{Generalization Barriers in Vision-Language-Action Models: Dataset, Architecture, and Processing Techniques}

\subsubsection{Differences Between Training Data and Procgen}

\textbf{Dataset Domain Discrepancies:} A fundamental obstacle limiting model performance arises from substantial differences between the training datasets and the Procgen evaluation dataset. Procgen environments are procedurally generated, Atari-like 2D games designed to evaluate visual and motor skills of reinforcement learning (RL) agents. Each subdataset within Procgen exhibits diverse environment layouts, tasks, objectives, reward structures, and discrete action spaces, predominantly involving directional movements and specific game-related interactions.

Conversely, the training datasets vary significantly across models:

GPT 4x models were primarily trained on expansive web-scale vision-language data, lacking a dedicated focus on action-oriented interactions.

OpenVLA is a vision-language-action model specifically fine-tuned on approximately 970,000 robotics demonstrations from the Open X-Embodiment dataset \cite{openx2024}. These demonstrations are strictly limited to continuous robotic manipulation tasks recorded from third-person camera views.

Pi0 models also leverage robotics trajectory data featuring continuous action spaces. Their training incorporates a subset of OpenX-Embodiment alongside a private dataset containing approximately 903 million timesteps, significantly sourced from single and dual-arm robotic manipulation scenarios \cite{pi02024}.

The substantial domain gap—real-world robotic manipulation versus simulated game interaction, continuous versus discrete action spaces, and physical manipulation versus simple directional controls—makes zero-shot adaptation particularly challenging for vision-language-action (VLA) models.

\textbf{Action Space Awareness:} Each model exhibits varying degrees of awareness regarding the action space constraints:

GPT 4x models are explicitly informed through the provided prompts about valid action values and corresponding verbal descriptions.

OpenVLA employs output clamping, effectively limiting predictions strictly within valid action ranges upon unnormalization based on dataset statistics.

Pi0 Base and Pi0 Fast rely on training-induced normalization but lack explicit clamping, resulting in a higher chance of invalid predictions outside permissible action ranges.

\textbf{Proprioceptive State Handling:} Procgen datasets lack proprioceptive state information, a critical input for Pi0 models during training. To compensate, zero arrays replace missing proprioceptive states during inference, negatively impacting Pi0 model performance by introducing unnatural, non-representative states. While OpenVLA optionally accepts proprioceptive states, the absence of these in Procgen data inherently limits the richness of information available for inference.

\textbf{Image Input Variability:} Significant discrepancies in image inputs further exacerbate model adaptation difficulties:

Resolution mismatch: Training images for OpenVLA and Pi0 models are at 224x224 resolution, whereas Procgen images are limited to 64x64 pixels, necessitating zero-padding and resizing, likely impairing visual perception and subsequent performance.

View differences: OpenVLA expects a single image input, matching the Procgen input structure, potentially enhancing its adaptability. In contrast, Pi0 models trained on multiple views suffer from performance degradation due to missing additional viewpoints being substituted by zero arrays.

\subsubsection{Difficulty Adapting to Increased Image Complexity}

We quantified image complexity through Shannon entropy \cite{shannon} and Delentropy \cite{delentropy2020} across 80 episodes over 16 datasets as seen in Figures \ref{fig:DatasetsByEntropy}, \ref{fig:fig19}, and \ref{fig:fig20}, revealing distinctive performance patterns. The formula used to calculate the entropies can be seen in Appendix section \ref{entropymetrics}:
\begin{figure}[h]
    \centering
    \includegraphics[width=1\linewidth]{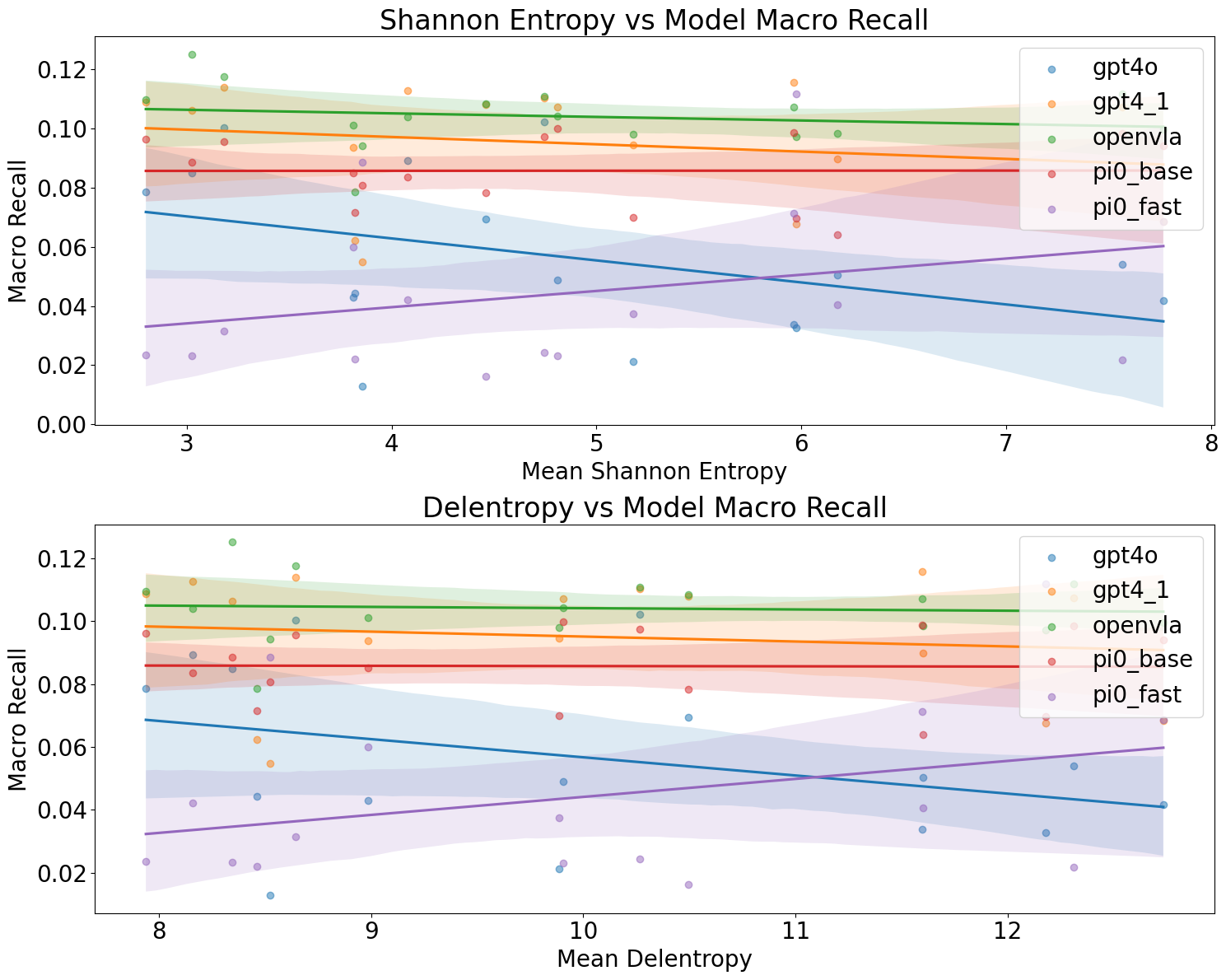}
    \caption{Entropy measures VS Model Macro Recall.  }
    \label{fig:fig19}
\end{figure}

\begin{figure}
    \centering
    \includegraphics[width=1\linewidth]{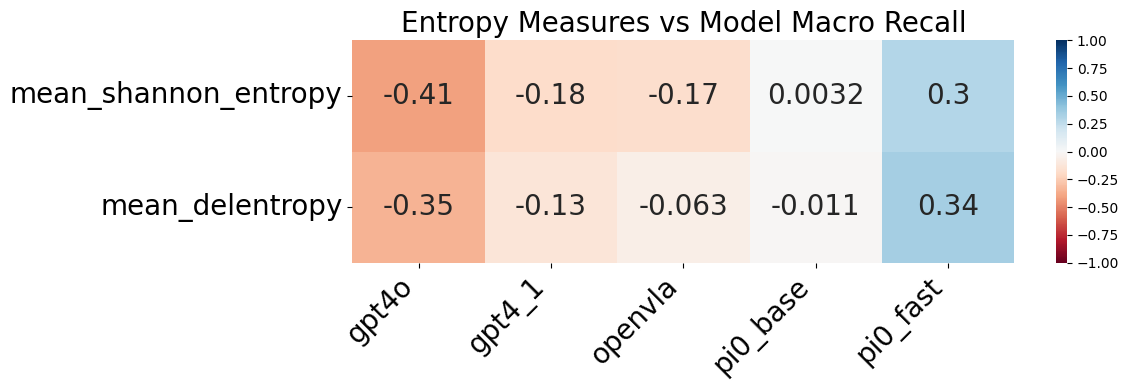}
    \caption{Correlation matrix for Entropy measures VS Model Macro Recall. }
    \label{fig:fig20}
\end{figure}

GPT-4 models (4o and 4.1) demonstrated moderate to weak negative correlations (Shannon entropy: -0.409 and -0.183; Delentropy: -0.345 and -0.126, respectively), suggesting poorer performance with increasing complexity.

OpenVLA and Pi0 Base exhibited negligible correlations with entropy metrics (OpenVLA: -0.173 Shannon, -0.063 Delentropy; Pi0 Base: 0.003 Shannon, -0.011 Delentropy), implying relative independence from image complexity.

Pi0 Fast uniquely displayed moderate positive correlations (Shannon and Delentropy: 0.296 and 0.336), indicating improved performance with higher image complexity.

These findings highlight that image complexity impacts GPT-4 models negatively, benefits Pi0 Fast, and minimally affects OpenVLA and Pi0 Base. The moderate correlation strength indicates complexity alone does not fully explain model performance variability.

\subsubsection{Impact of VLA Training, Decoding, and Output Processing Techniques}

Distinct training, decoding, and output processing strategies significantly shape model generalization capabilities:

OpenVLA utilizes a robust output processing approach, training to output actions within a normalized range of [-1,1], with strict clamping enforced. Consequently, its predictions, upon unnormalization, consistently map within valid action spaces, eliminating invalid outputs and markedly enhancing zero-shot generalization.

Pi0 Base also targets normalized outputs within [-1,1] without enforced clamping. Its diffusion-based architecture inherently predicts actions ``close'' to familiar training distributions. This approach yields relatively fewer invalid outputs and moderately effective generalization compared to purely autoregressive models, but remains inferior to OpenVLA.

Pi0 FAST, while similarly trained on normalized action outputs, employs autoregressive decoding without clamping, lacking the intrinsic action proximity measure found in diffusion models. Consequently, Pi0 FAST predictions deviate significantly from valid actions when generalization fails, resulting in extremely high invalid output rates and poor overall performance, except for moderate macro precision.

Collectively, these findings underscore critical architectural and training considerations necessary for enhancing generalization across diverse and unseen discrete action environments.

\section{Conclusion}

In this study, we systematically evaluated the generalization capabilities of contemporary VLAs and VLMs on procedurally generated discrete-action environments from the Procgen dataset in a zero-shot setting. Our analysis highlighted significant limitations arising from architectural constraints, training paradigms, and input-output biases inherent to the models. The stark domain discrepancy between training data—primarily continuous-action robotics datasets, and general web-scale vision language data—and discrete-action game environments proved to be a critical barrier to effective zero-shot generalization.

We identified notable differences in model behaviors linked directly to their architectures, training strategies, and input/output processing techniques. OpenVLA's robust action-space clamping technique consistently provided superior generalization, minimizing invalid outputs and exhibiting relative resilience to out-of-distribution scenarios. Conversely, autoregressive models like GPT-4x displayed substantial difficulty in generalizing, especially under complex image conditions, and frequently defaulted to idle or biased action choices. Additionally, Pi0 models showed intermediate performance influenced heavily by their diffusion-based (Pi0 Base) or autoregressive (Pi0 FAST) decoding methods, with Pi0 FAST being notably sensitive to image complexity and unable to restrict the majority of its predictions to a desired output range.

Our findings underscore the necessity for architectural innovations, refined training methodologies, and enhanced output processing techniques to bridge the gap between diverse action domains. Future research should prioritize developing more generalized training datasets that better reflect the variety of potential application environments, alongside methods to adaptively handle different forms of action representations. These advancements hold promise for enabling VLAs and VLMs to operate effectively across an increasingly diverse and unpredictable range of real-world tasks.

\bibliographystyle{plain}  
\bibliography{LLM.bib}  

\newpage
\section{Appendix}
\subsection{Heatmap of Model Macro Recall Scores on 16 Subdatasets}
\begin{figure}[H]
    \centering
    \includegraphics[width=1\linewidth]{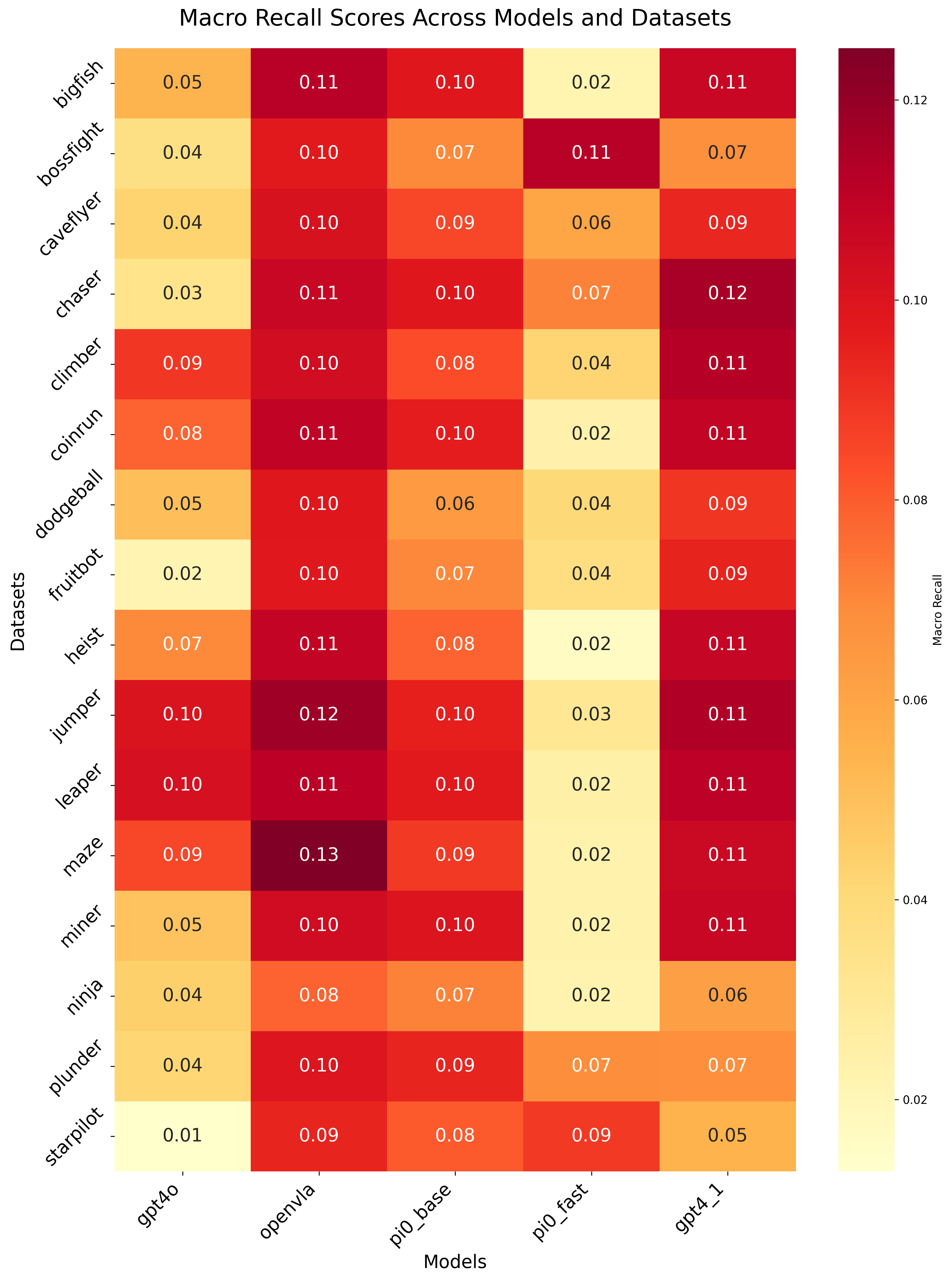}
    \caption{Heatmap of the 5 models' the macro recall scores over 16 subdatasets. OpenVLA has consistently good macro recall over all subdatasets. Datasets such as ninja with complex action space are difficult across all 5 models.}
    \label{fig:model_heatmap}
\end{figure}

\newpage
\subsection{Individual subdataset confusion matrices}
\label{appendix_confusionmatrix}

\begin{figure}[H]
	\centering
	\includegraphics[width=9cm]{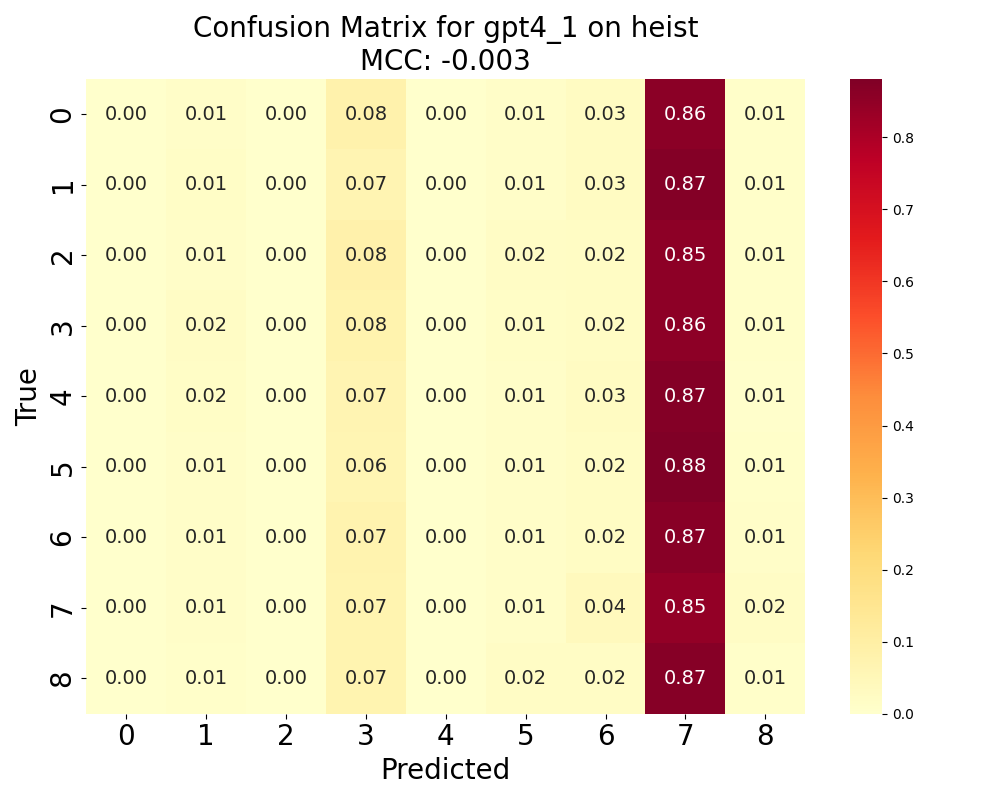}
	\caption{Confusion matrix for GPT 4.1 on Heist}
	\label{fig:fig21}
\end{figure}

\begin{figure}[h]
	\centering
	\includegraphics[width=9cm]{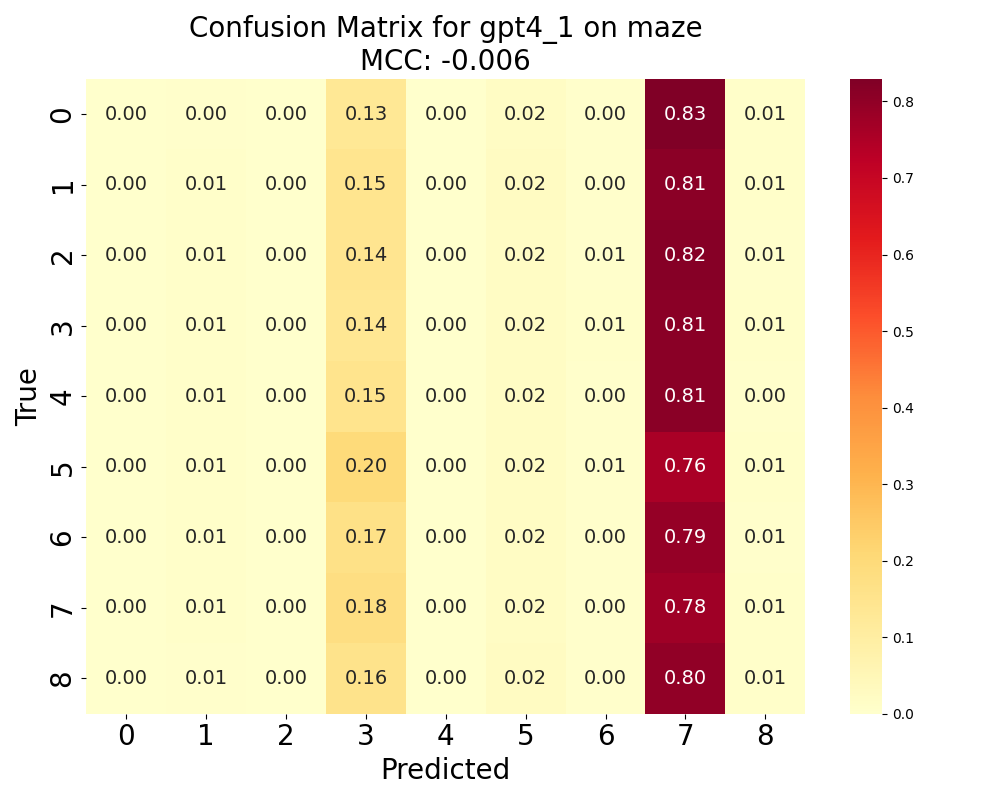}
	\caption{Confusion matrix for GPT 4.1 on Maze}
	\label{fig:fig22}
\end{figure}

\begin{figure}[h]
	\centering
	\includegraphics[width=9cm]{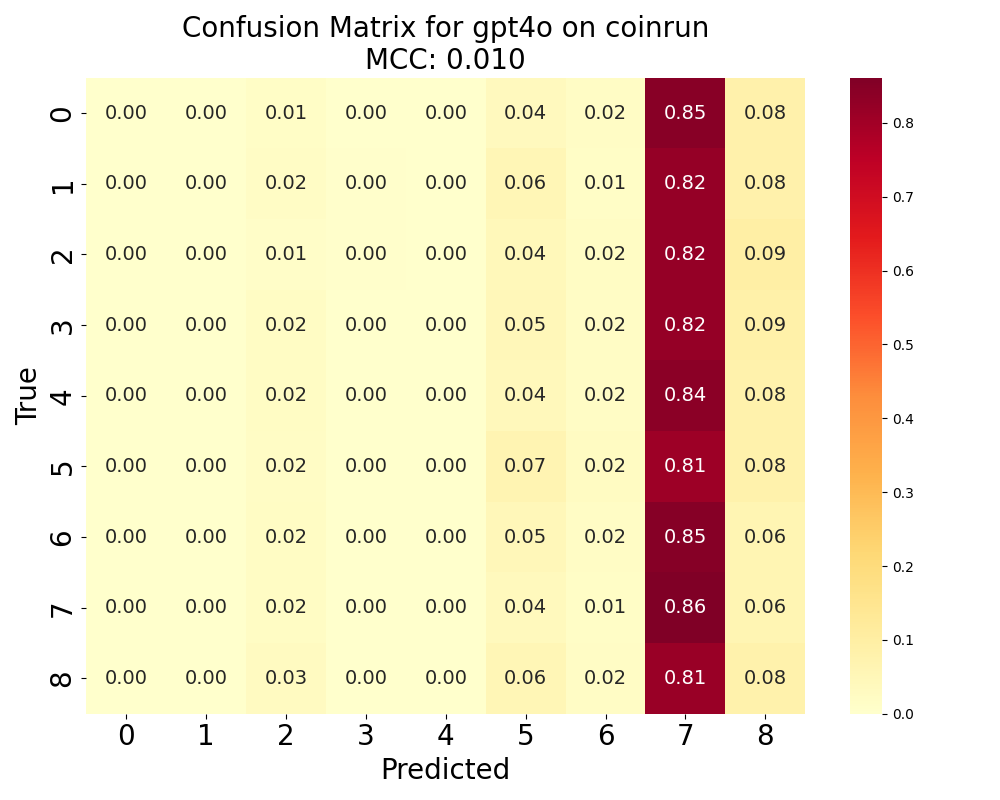}
	\caption{Confusion matrix for GPT 4o on Coinrun}
	\label{fig:fig23}
\end{figure}

\begin{figure}[h]
	\centering
	\includegraphics[width=9cm]{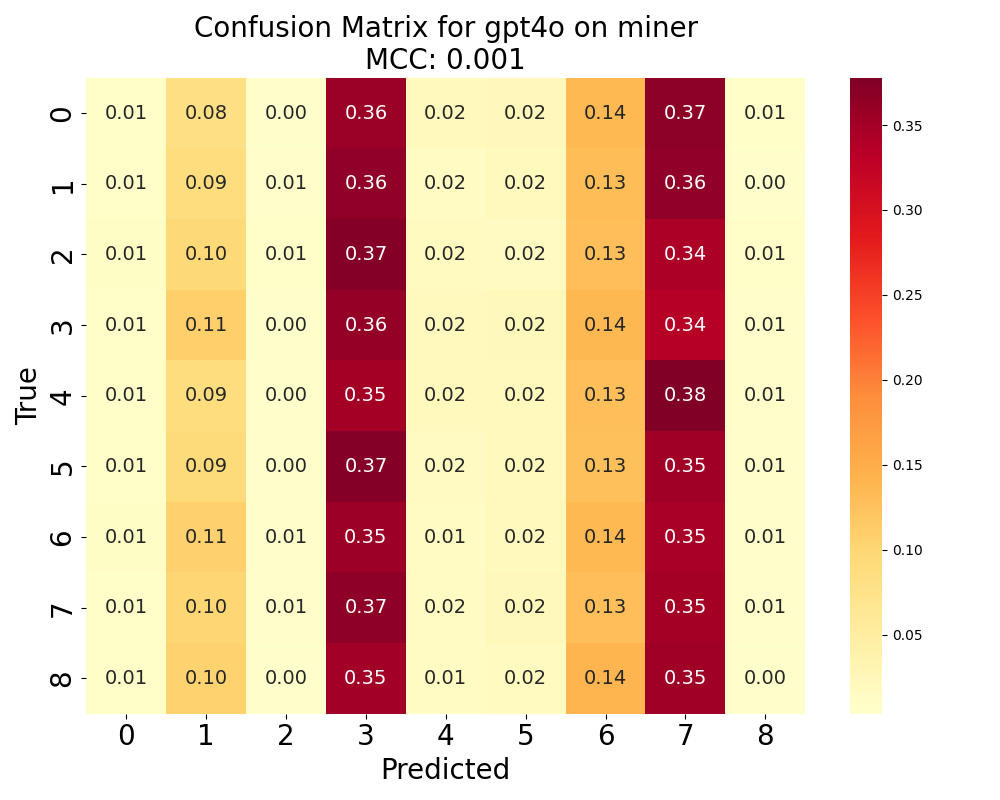}
	\caption{Confusion matrix for GPT 4o on Miner}
	\label{fig:fig24}
\end{figure}

\begin{figure}[h]
	\centering
	\includegraphics[width=9cm]{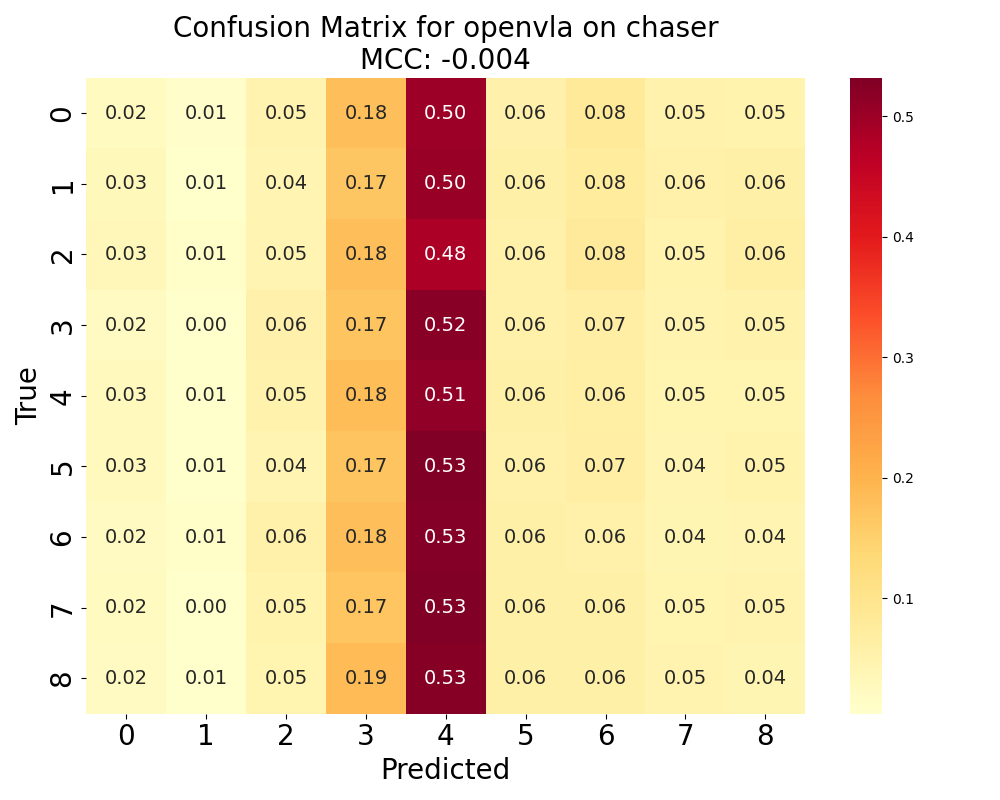}
	\caption{Confusion matrix for OpenVLA on Chaser}
	\label{fig:fig25}
\end{figure}

\begin{figure}[h]
	\centering
	\includegraphics[width=9cm]{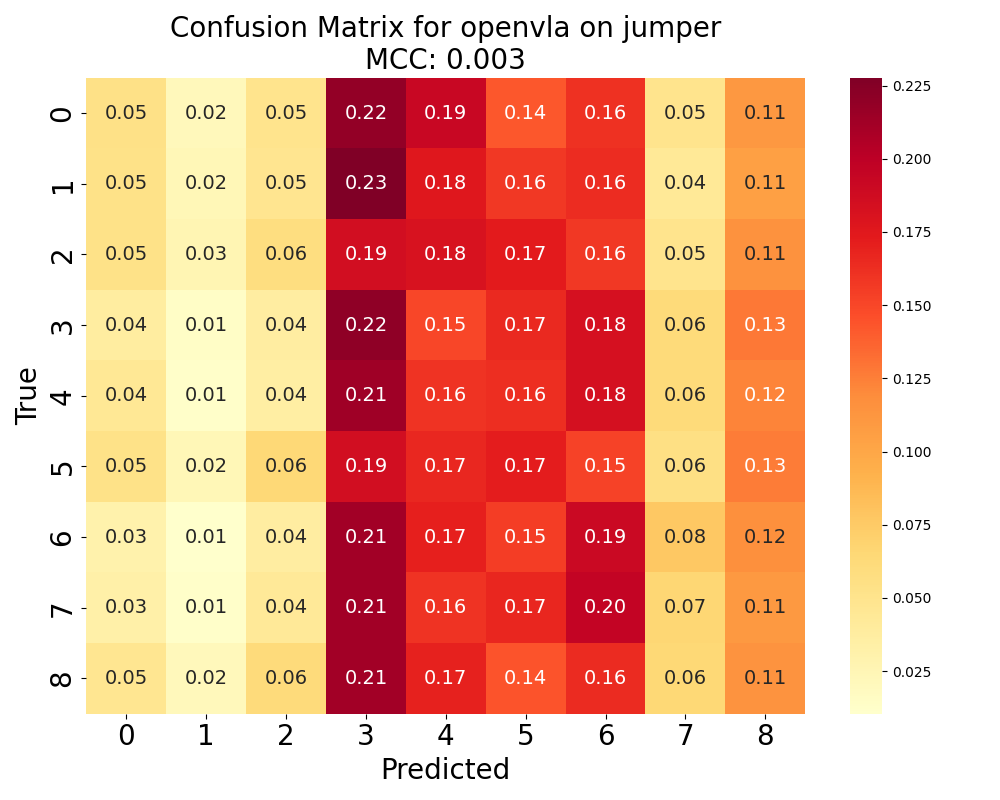}
	\caption{Confusion matrix for OpenVLA on Jumper}
	\label{fig:fig26}
\end{figure}

\begin{figure}[h]
	\centering
	\includegraphics[width=9cm]{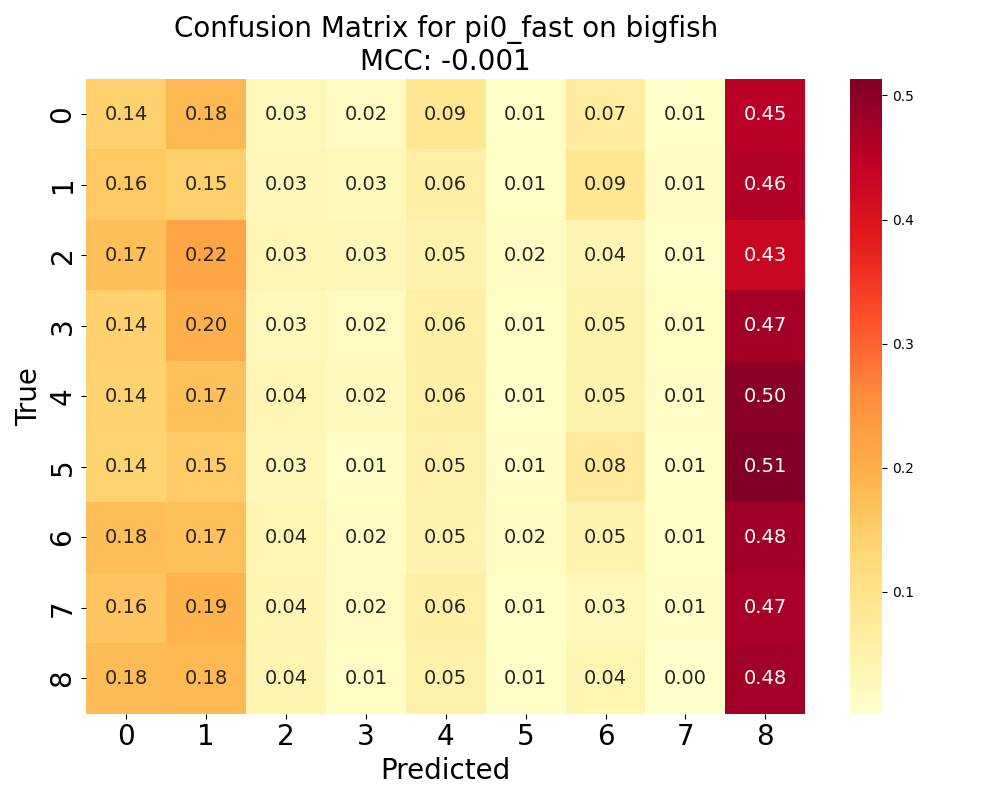}
	\caption{Confusion matrix for Pi0 FAST on Bigfish}
	\label{fig:fig27}
\end{figure}

\begin{figure}[h]
	\centering
	\includegraphics[width=9cm]{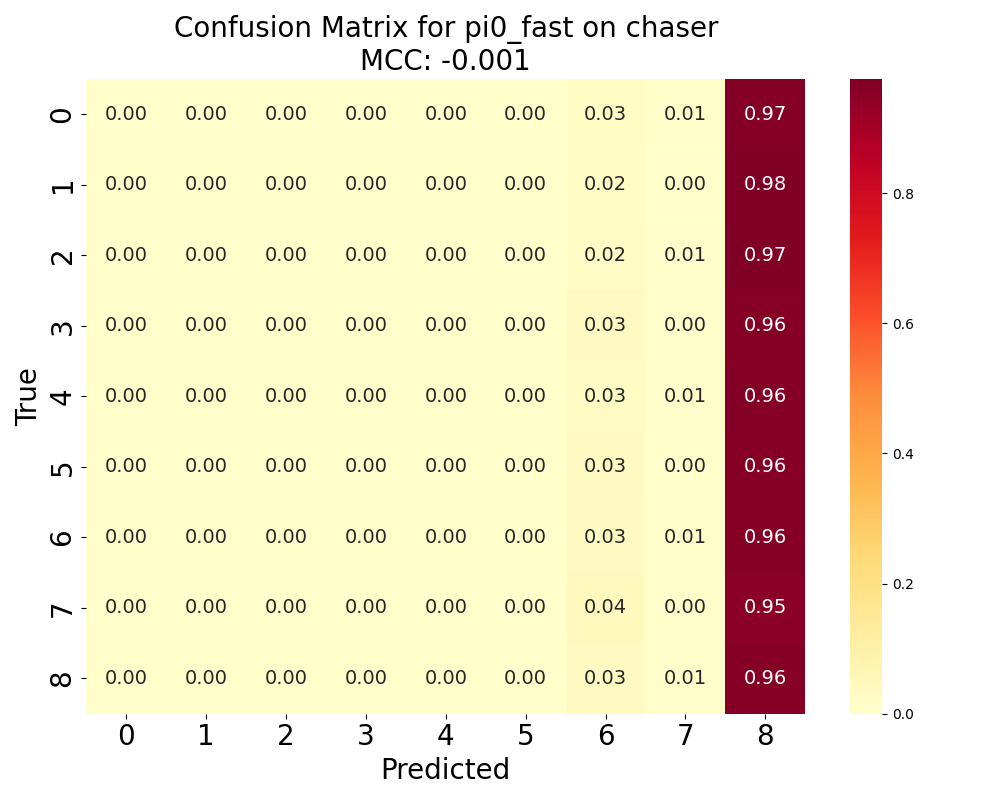}
	\caption{Confusion matrix for Pi0 FAST on Chaser}
	\label{fig:fig28}
\end{figure}

\begin{figure}[h]
	\centering
	\includegraphics[width=9cm]{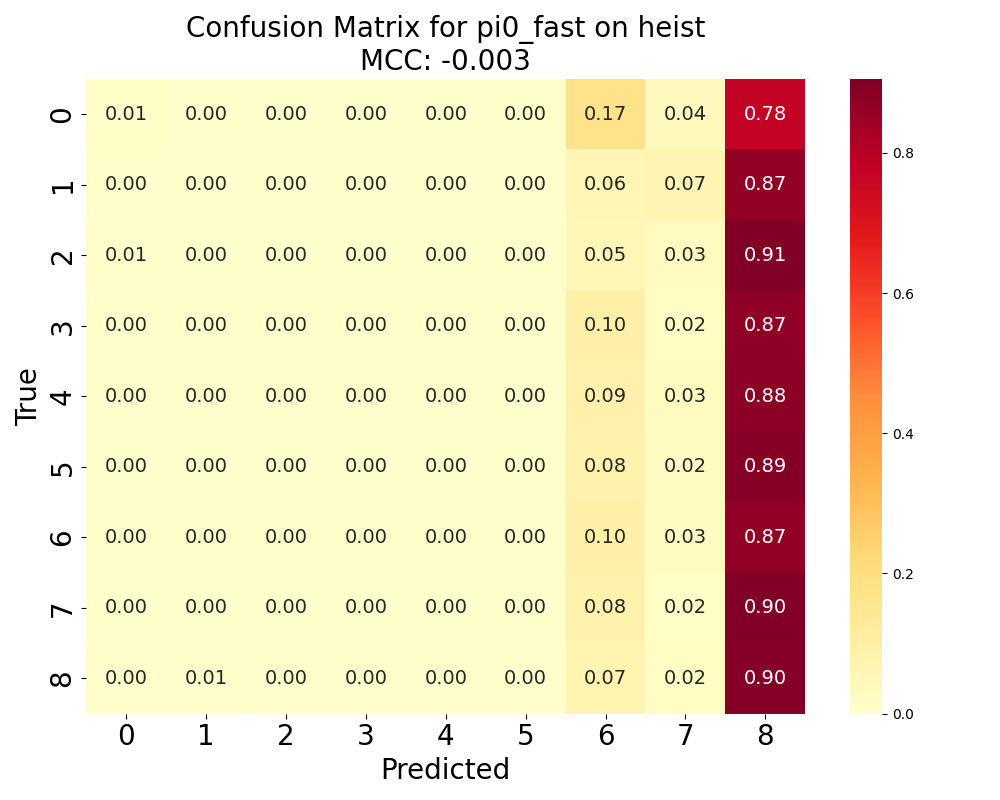}
	\caption{Confusion matrix for Pi0 FAST on Heist}
	\label{fig:fig29}
\end{figure}

\begin{figure}[h]
	\centering
	\includegraphics[width=9cm]{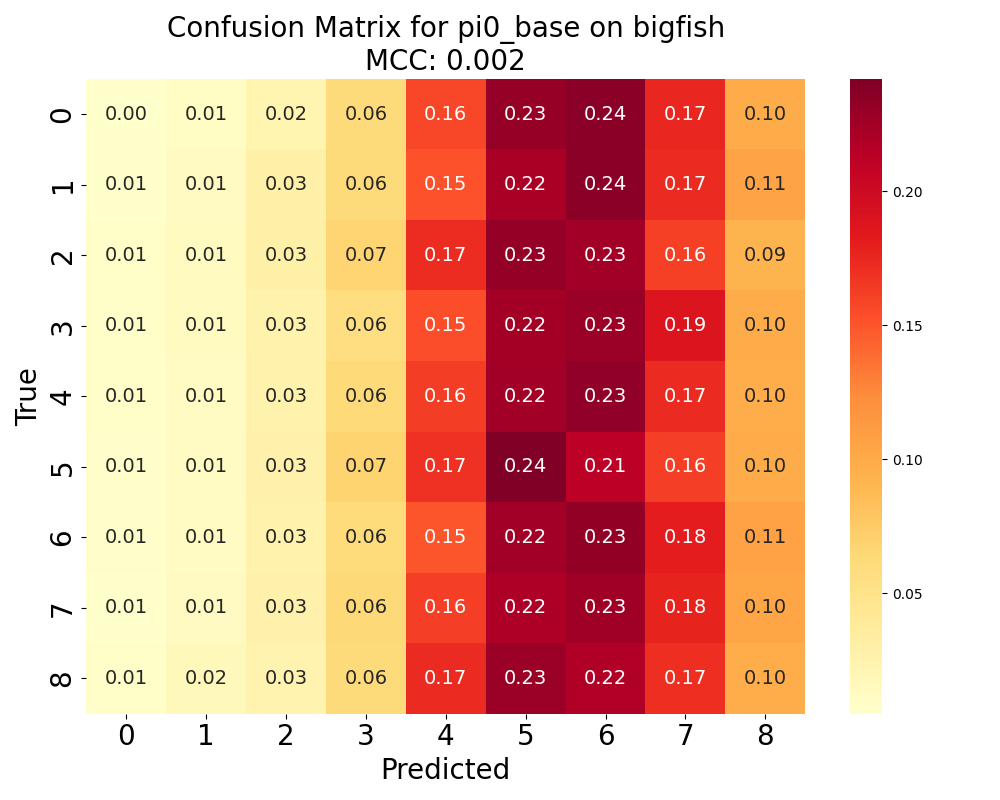}
	\caption{Confusion matrix for Pi0 Base on Bigfish}
	\label{fig:fig30}
\end{figure}

\begin{figure}[h]
	\centering
	\includegraphics[width=9cm]{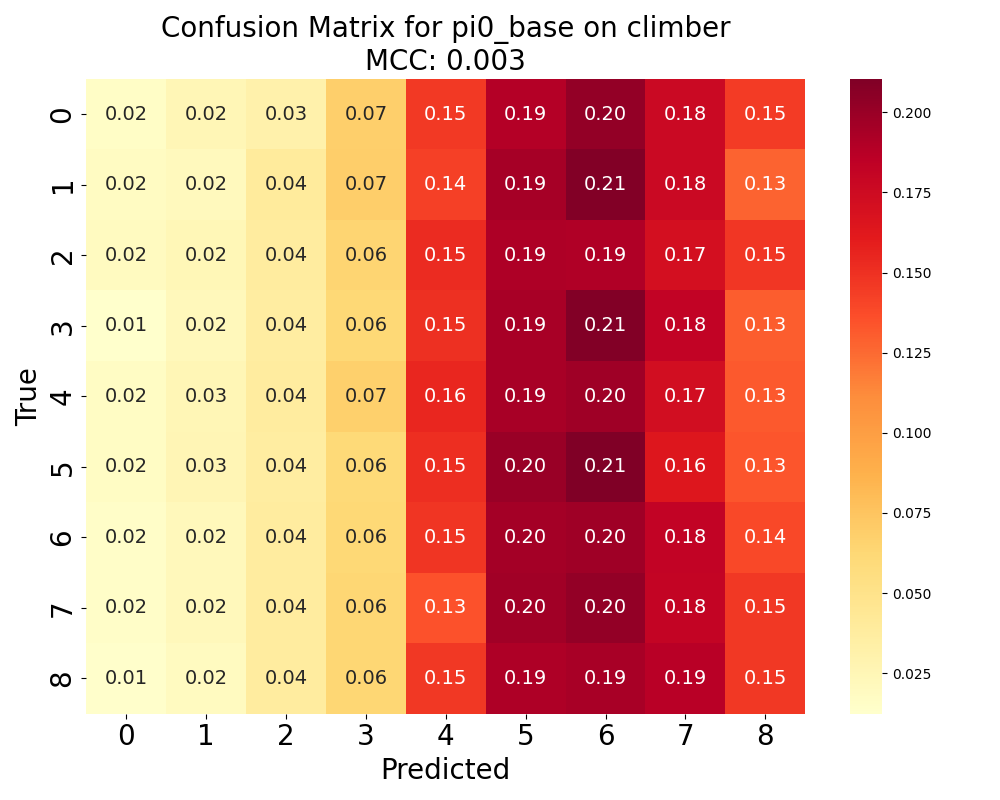}
	\caption{Confusion matrix for Pi0 Base on Climber}
	\label{fig:fig31}
\end{figure}

\begin{figure}[h]
	\centering
	\includegraphics[width=9cm]{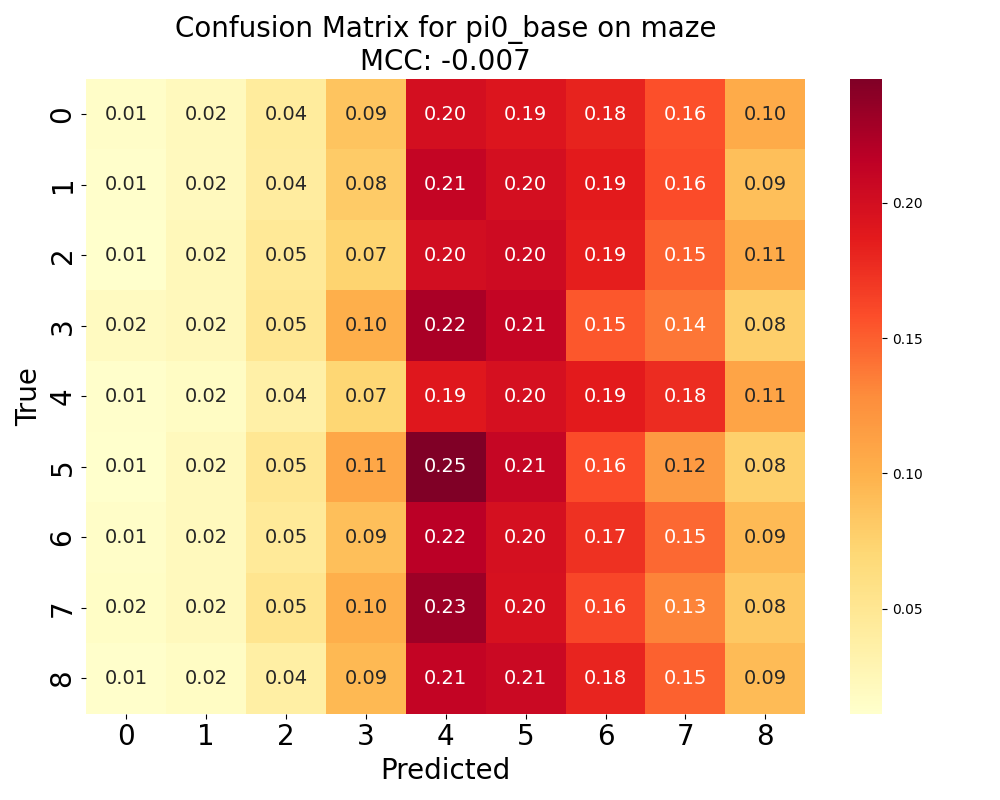}
	\caption{Confusion matrix for Pi0 Base on Maze}
	\label{fig:fig32}
\end{figure}

\begin{figure}[H]
    \centering
    \includegraphics[width=1\linewidth]{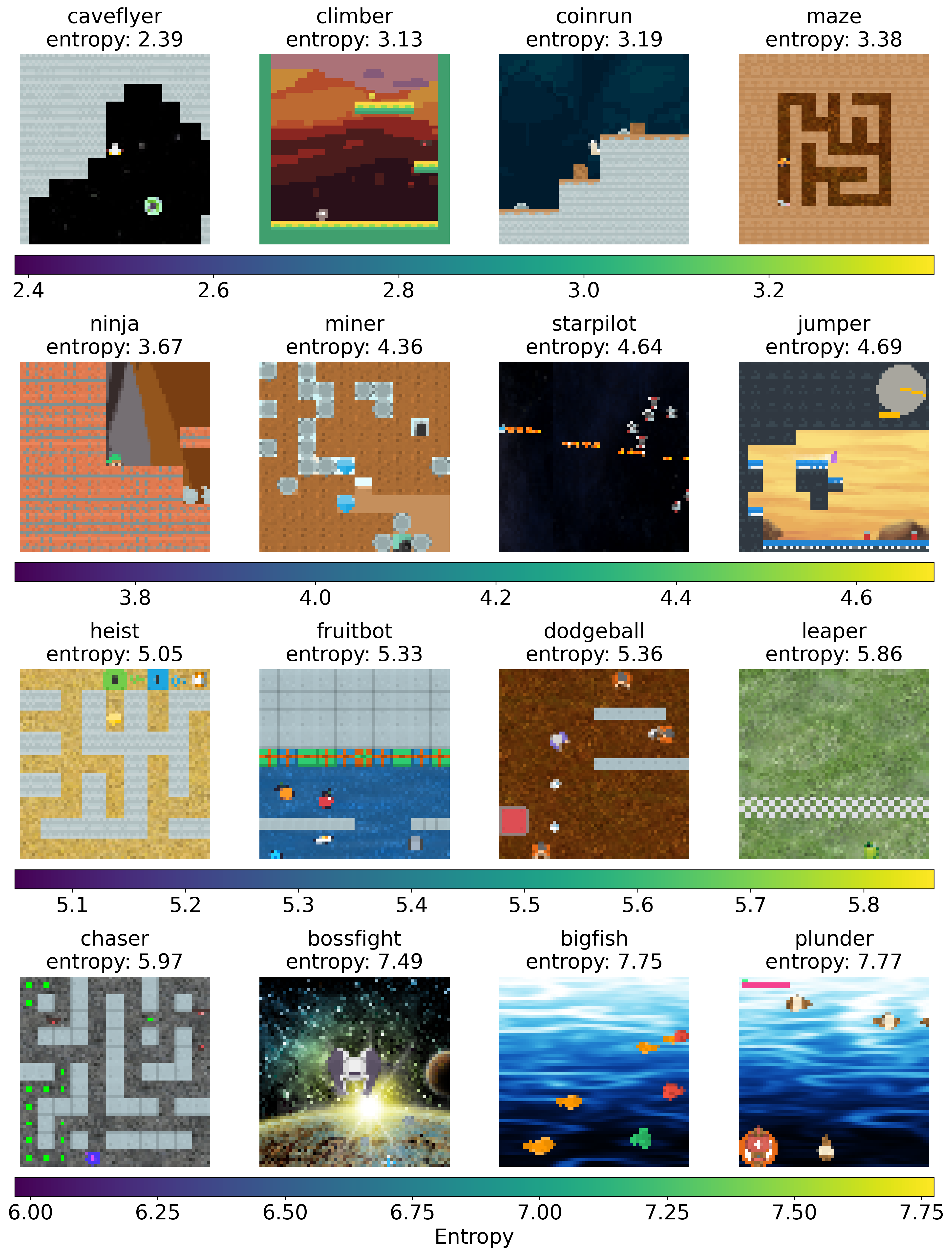}
    \caption{Datasets ranked by Shannon entropy}
    \label{fig:DatasetsByEntropy}
\end{figure}

Figures \ref{fig:fig21} to \ref{fig:fig32} are examples of confusion matrices of models on individual subdatasets to observe the pattern of action class collapse across models.

\FloatBarrier
\subsection{Entropy metrics}
\label{entropymetrics}
\begin{itemize}
    \item \textbf{Shannon Entropy}
    To study the complexity of the images in Procgen, we use Shannon Entropy, calculated over each subdataset's first 20 episodes. The standard Shannon entropy quantifies the amount of uncertainty or information content in an image by analyzing its pixel intensity distribution. The Shannon entropy of a grayscale image is defined as:
    
    \begin{equation}
        H = - \sum_{i=0}^{n-1} p_i \log p_i
    \end{equation}

    \item \textbf{Delentropy}
    Delentropy is a metric that leverages a probability density function, referred to as deldensity, which can be computed from image data. By analyzing how pixels and their spatial relationships co-occur within an image, deldensity enables delentropy to effectively characterize the underlying structural patterns present in the image. Delentropy is defined as:
    \begin{equation}
        H = - \sum_{i,j} p(f_x, f_y) \log_2 p(f_x, f_y)
    \end{equation}
\end{itemize}

\subsection{Datasets over Shannon entropy}

\subsection{Full entropy and model correlation matrix}
\begin{figure}[H]
	\centering
	\includegraphics[width=9cm]{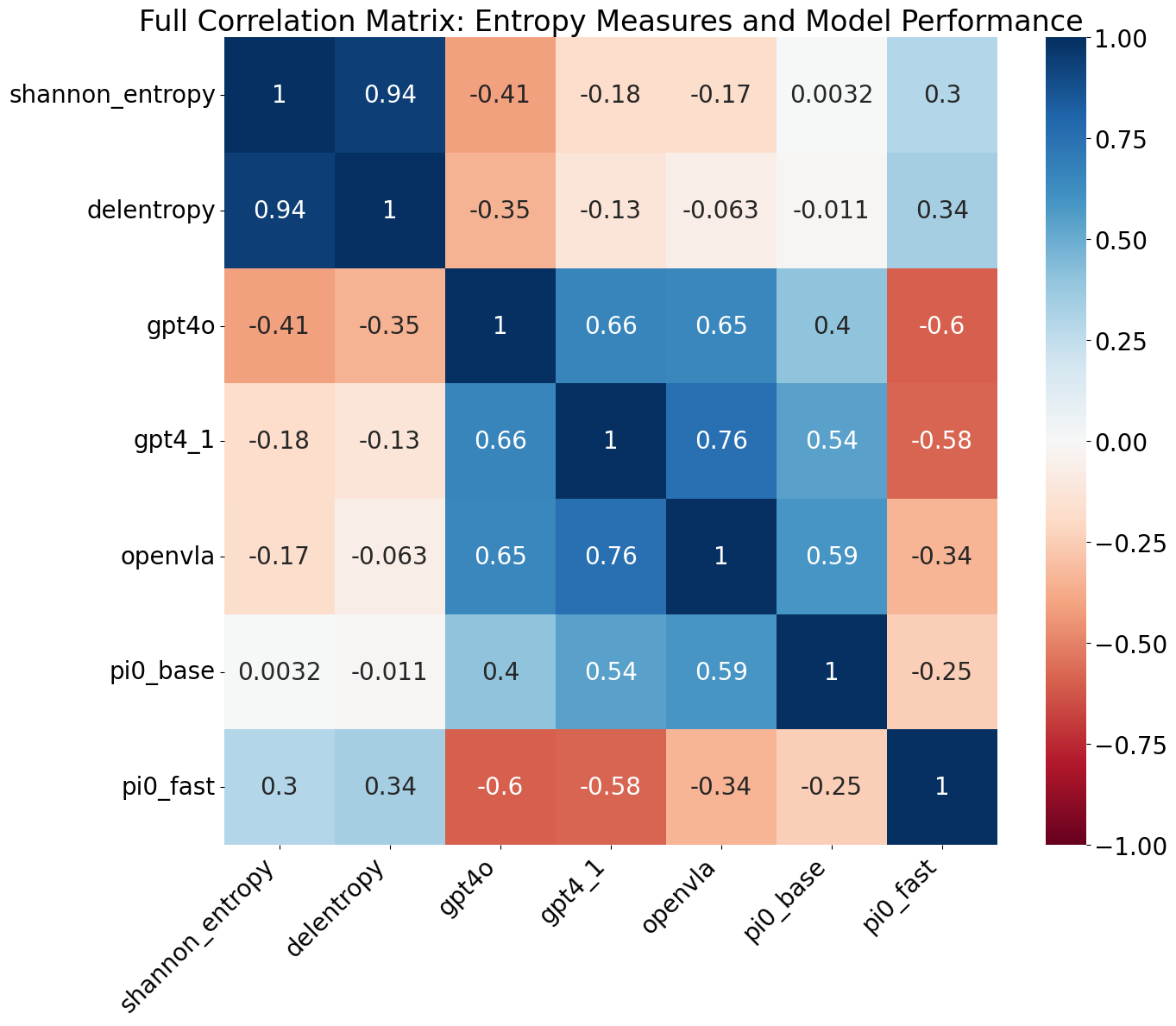}
	\caption{Correlation matrix for Entropy measures VS Model Macro Recall}
	\label{fig:fig33}
\end{figure}

\subsection{Genesis Framework}
\label{genesisprompts}
The Genesis framework consists of the following parts:
\begin{itemize}
    \item System-level Instructions: The overall goal, the constraints for inference, and the nature of the interaction (e.g., a simulated environment)
\end{itemize}
\begin{itemize}
    \item Task and Environment Context: Explicit descriptions of the specific Procgen sub-dataset task, its rules, and relevant environmental details
\end{itemize}
\begin{itemize}
    \item Multimodal Input Integration: The visual observation (current game state image) within the prompt structure
\end{itemize}
\begin{itemize}
    \item Action Space Definition: The available actions, their format, and their corresponding verbal descriptions
\end{itemize}
\begin{itemize}
    \item Output Instructions: The precise output structure such that the model handles the complexity of generating a valid set of probabilities for the actions in the action space
\end{itemize}

Although the information provided to GPT 4o and 4.1 was the same, the prompt constructed for GPT 4.1 was more rigorously curated than the one for GPT 4o. The prompt for GPT 4.1 employed more prompt engineering techniques to minimize the number of invalid outputs. The new prompt explicitly specifies the simulated and hypothetical nature of the dataset and capitalizes key instructions for emphasis.
Additionally, the new prompt tailors the input type descriptions strictly to those encountered in the current dataset, rather than stating all possible types of input from the broader collection of Multinet datasets.
Here are examples of the old and new prompts constructed by Genesis for a given timestep of the subdataset Heist:

\begin{itemize}
    \item The old prompt, used to profile GPT4o:
    ``\textit{You are an AI agent to solve the task called \textbf{steal the gem}. In this environment: Collect keys of different colors. Open colored locks. Reach and collect the hidden gem. You should produce a proper action output to achieve the final goal given the current progress so far given the current state information. The current state can be any forms, such as images, continuous/discrete vectors, or texts. The actions available: A discrete action has the available options as key-value pairs, {Option index}: {Option description}.\\0. Agent movement action =$>$ Discrete. Options: {0: 'LEFT + DOWN', 1: 'LEFT', 2: 'LEFT + UP', 3: 'DOWN', 4: 'Do Nothing', 5: 'UP', 6: 'RIGHT + DOWN', 7: 'RIGHT', 8: 'RIGHT + UP'}.\\ You must generate your output keeping the following format: A list starting with '[' and ending with ']'. Each position corresponds to each action index. Each position in that list should be a hashmap starting with '{' and ending with '}'. The hashmap should contain a key for each option index of that action, and the value for that key corresponds to the probability that this option should be selected as the next step. All probabilities across all actions, as opposed to per action or hashmap, must sum up to 1.0. You should not include any other words or characters in your response. }''

    \item The new prompt, used to profile GPT 4.1:
    ``\textit{We are running a simulation for an AI agent playing a video game. Your role is to evaluate potential moves based on a snapshot. The description of this hypothetical scenario is \textbf{steal the gem}. In this environment: Collect keys of different colors. Open colored locks. Reach and collect the hidden gem. You should produce a proper action output to achieve the final goal given the current progress so far given the current state information. The current state consists of an image, which is a snapshot of the game screen, and a text description of the objective. The actions available: A discrete action has the available options as key-value pairs, {Option index}: {Option description}.
    \\0. Agent movement action =$>$ Discrete. Options: {0: 'LEFT + DOWN', 1: 'LEFT', 2: 'LEFT + UP', 3: 'DOWN', 4: 'Do Nothing', 5: 'UP', 6: 'RIGHT + DOWN', 7: 'RIGHT', 8: 'RIGHT + UP'}.\\
    You MUST generate your output keeping the following format: A list starting with '[' and ending with ']'. Each position corresponds to each action index. Each position MUST be a hashmap starting with '{' and ending with '}'. The hashmap should contain a key for each option index of that action, and the value for that key corresponds to the probability that this option should be selected as the next step. ALL probabilities across all actions, as opposed to per action or hashmap, MUST sum up to 1.0. You should not include any other words or characters in your response. }''
\end{itemize}

\end{document}